\documentclass[applsci,article,moreauthors,pdftex]{mdpi} 
\usepackage[utf8]{inputenc}
\usepackage{lipsum,microtype}
\usepackage[labelformat=simple]{subfig}

\usepackage{framed,multirow}
\usepackage{latexsym}
\usepackage[utf8]{inputenc}
\usepackage[T1]{fontenc}
\usepackage{graphicx,float}
\usepackage{cite}
\usepackage{amsmath,amssymb,amsfonts}
\usepackage{algorithmic}
\setitemize{parsep=6pt,itemsep=0pt,leftmargin=*,labelsep=5.5mm}
\setenumerate{parsep=6pt,itemsep=0pt,leftmargin=*,labelsep=5.5mm}
\setlist[description]{itemsep=0mm}

\usepackage{lettrine}
\usepackage[utf8]{inputenc}
\usepackage{hyperref}
\firstpage{1} 
\pubvolume{00}
\issuenum{1}
\articlenumber{5}
\pubyear{2020}
\copyrightyear{2020}
\history{Received: 16 September 2020; Accepted: 20 October 2020; Published: date}

\continuouspages{yes}
\updates{yes}
\pdfoutput=1

\Title{DeepSOCIAL: Social Distancing Monitoring and Infection Risk Assessment in COVID-19 Pandemic}

\Author{Mahdi Rezaei 
 $^{1, *, \dagger}$\orcidA{} and Mohsen Azarmi $^{2, \dagger}$\orcidB{}}

\address{$^{1}$ \quad Institute for Transport Studies, The University of Leeds, 34-40 University Road, Leeds, LS2 9JT, UK
\\
$^{2}$ \quad Faculty of Computer and IT Engineering, Qazvin Islamic Azad University, Qazvin, 15195-34199, IR}   

\corres{Correspondence: m.rezaei@leeds.ac.uk; Tel.: +44-113-3435342}

\firstnote{The authors contributed~equally.}

\abstract{Social distancing is a recommended solution by the World Health Organisation (WHO) to minimise the spread of COVID-19 in public places. The majority of governments and national health authorities have set the 2-meter physical distancing as a mandatory safety measure in shopping centres, schools and other covered areas. In this research, we develop a hybrid \textit{Computer Vision} and YOLOv4-based \textit{Deep Neural Network} (DNN) model for automated people detection in the crowd in indoor and outdoor environments using common CCTV security cameras. The proposed DNN model in combination with an adapted inverse perspective mapping (IPM) technique and SORT tracking algorithm leads to a robust people detection and social distancing monitoring. The model has been trained against two most comprehensive datasets by the time of the research---the Microsoft Common Objects in Context (MS COCO) and Google Open Image datasets. The system has been evaluated against the Oxford Town Centre dataset (including 150,000 instances of people detection) with superior performance compared to three state-of-the-art methods. The evaluation has been conducted in challenging conditions, including occlusion, partial visibility, and under lighting variations with the mean average precision of 99.8\% and the real-time speed of 24.1 fps. We also provide an online infection risk assessment scheme by statistical analysis of the spatio-temporal data from people's moving trajectories and the rate of social distancing violations.
We identify high-risk zones with the highest possibility of virus spread and infection. This may help authorities to redesign the layout of a public place or to take precaution actions to mitigate high-risk zones. The developed model is a generic and accurate people detection and tracking solution that can be applied in many other fields such as autonomous vehicles, human action recognition, anomaly detection, sports, crowd analysis, or any other research areas where the human detection is in the centre of attention.}

\keyword{social distancing; COVID-19; human detection and tracking; distance estimation; deep~convolutional neural networks; crowd monitoring; pedestrian detection; inverse perspective mapping.}

\begin{document}

\section{Introduction}\label{Introduction}
The novel generation of the coronavirus disease (COVID-19) was reported in late December 2019 in Wuhan, China. After~only a few months, the~virus became a global outbreak in 2020. On~May 2020 The World Health Organisation (WHO) announced the situation as pandemic \citep{who2020a,who2020b}. The~statistics by WHO on 8th October 2020 confirm 36 million infected people and a scary number of 1,056,000 deaths in 200~countries.

With the growing trend of patients, there is still no effective cure or available treatment for the virus. While scientists,  healthcare organisations, and~researchers are continuously working to produce appropriate medications or vaccines for the deadly virus, no definite success has been reported at the time of this research, and~there is no certain treatment or recommendation to prevent or cure this new disease. Therefore, precautions are taken by the whole world to limit the spread of infection. These~harsh conditions have forced the global communities to look for alternative ways to reduce the spread of the~virus. 

\begin{figure}[t!]
\centering
\begin{tabular}{cc}

\includegraphics[width=0.48\linewidth]{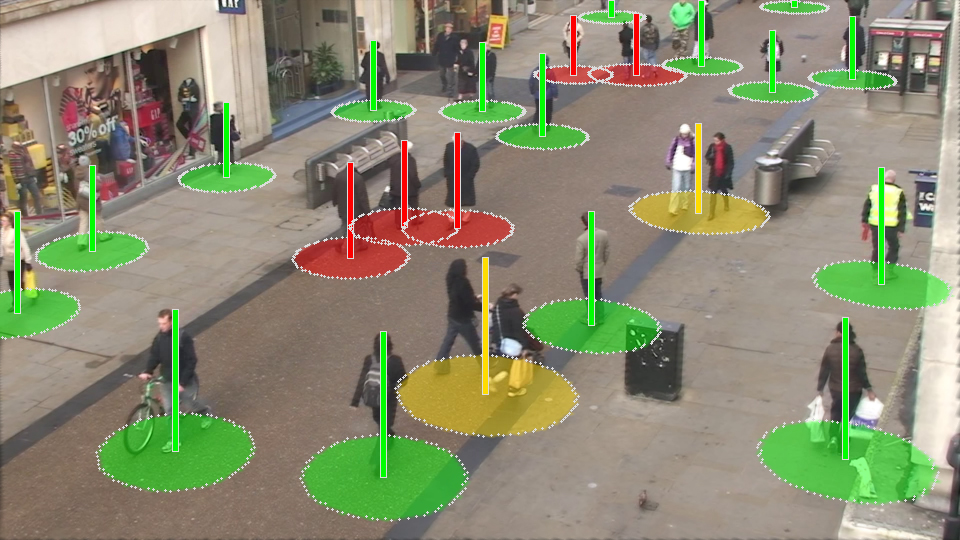} & \includegraphics[width=0.48\linewidth]{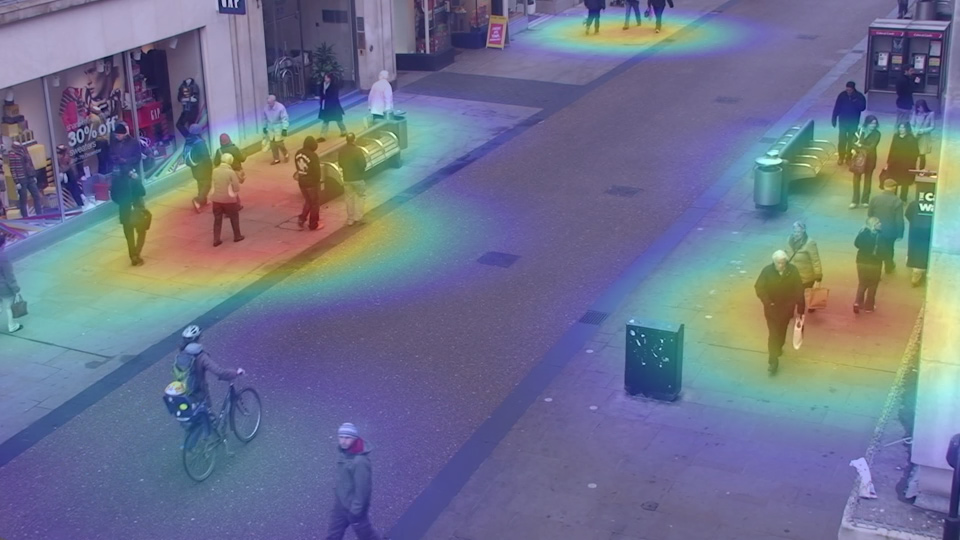} \\
(\textbf{a}) Social distancing monitoring & (\textbf{b}) Accumulated infection risk (red zones) \\
& due to breaches of the social-distancing
\end{tabular}

%
\caption{People detection, tracking, and~risk assessment in Oxford Town Centre, using a public CCTV~camera.}
\label{teaser}
\end{figure}

Social distancing, as~shown in Figure~\ref{teaser}a, refers to precaution actions to prevent the proliferation of the disease, by~minimising the proximity of human physical contacts in covered or crowded public places (e.g., schools, workplaces, gyms, lecture theatres, etc.) to stop the widespread accumulation of the infection risk (Figure \ref{teaser}b). 

For several months, the~World Health Organisation believed that COVID-19 was only transmittable via droplets emitted when people sneeze or cough and the virus does not linger in the air. However, on~8 July 2020, the~WHO announced:

\begin{quote}
\textit{``There is emerging evidence that COVID-19 is an airborne disease that can be spread by tiny particles suspended in the air after people talk or breathe, especially in crowded, closed environments or poorly ventilated settings''}~\citep{who2020b}.
\end{quote}

Therefore, social distancing now claims to be even more important than thought before, and~one of the best ways to stop the spread of the disease in addition to wearing face masks. Almost all countries are now considering it as a mandatory~practice.

According to the defined requirements by the WHO, the~minimum distance between individuals must be at least 6 feet (1.8 m) in order to observe an adequate social distancing among the people~\citep{hensley2020a}.

Recent research has confirmed that people with mild or no symptoms may also be carriers of the novel coronavirus infection \citep{ecdpc2020}. Therefore, it is important all individuals maintain controlled behaviours and observe social distancing. 
Many research works such as \citep{ferguson2006,thu2020effect,morato2020optimal} have proved social-distancing as an effective non-pharmacological approach and an important inhibitor for limiting the transmission of contagious diseases such as H1N1, SARS, and~COVID-19.

\begin{figure}[t!]
\centering
\includegraphics[width=0.6\linewidth]{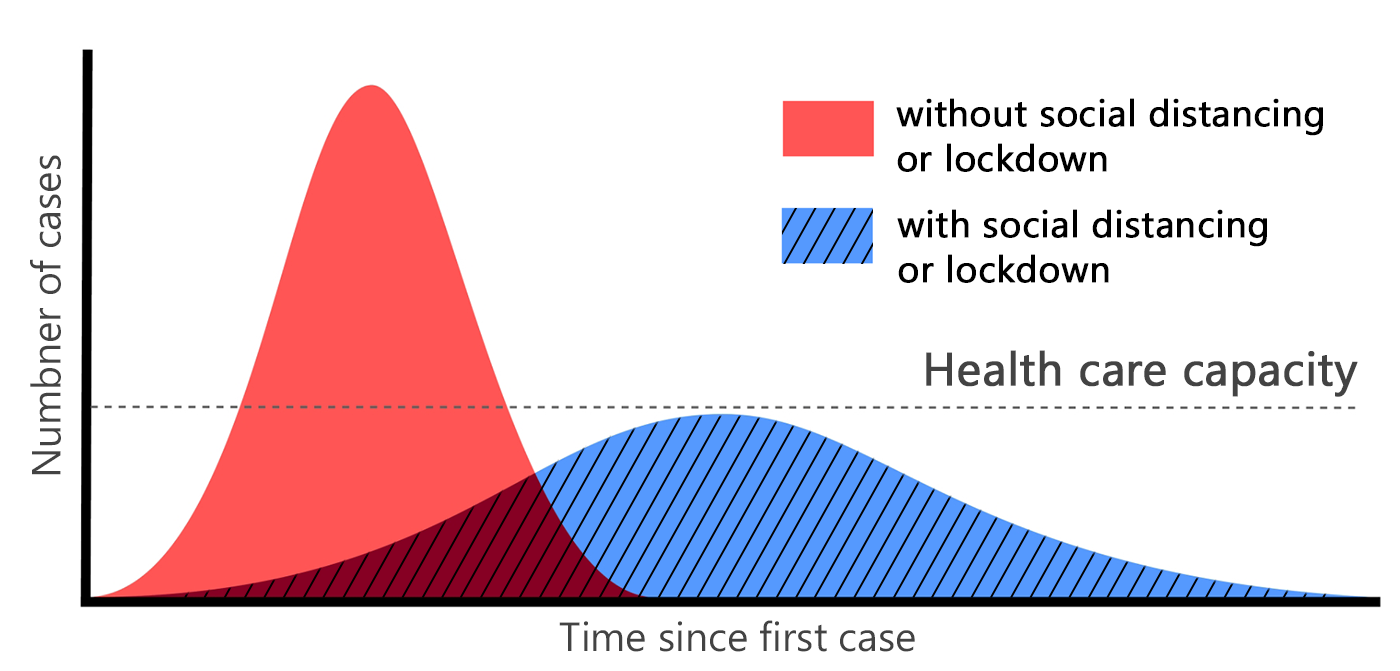}
\caption{Gaussian distribution of infection transmission rate for a given population, with~and without social distancing~obligation.}
\label{covid19}
\end{figure}

Figure~\ref{covid19} demonstrates the effect of following appropriate social distancing guidelines to reduce the rate of infection transmission among individuals \citep{fong2020a,ahmed2018a}. 
A wider Gaussian curve with a shorter spike within the range of the health system service capacity makes it easier for patients to fight the virus by receiving continuous and timely support from the health care organisations. Any unexpected sharp spike and rapid infection rate (such as the red curve in Figure~\ref{covid19}), will lead to service failure, and~consequently, exponential growth in the number of~fatalities. 

During the COVID-19 pandemic, governments have tried to implement a variety of social distancing practices, such as restricting travels, controlling borders, closing pubs and bars, and~alerting the society to maintain a distance of 1.6 to 2 m from each other \citep{australian2020}. However, monitoring the amount of infection spread and efficiency of the constraints is not an easy task. People require to go out for essential needs such as food, health care and other necessary tasks and jobs. 
Therefore, many other technology-based solutions such as \citep{nguyen2020enabling,punn2020next} and AI related research such as \citep{shi2020review,R0-2020,punn2020monitoring} have tried to step in to help the health and medical community in copping with COVID-19 challenges and successful social distancing practices. These works vary from GPS-based patient localisation and tracking to segmentation, and~crowd~monitoring.

In such situations, Artificial Intelligence can play an important role in facilitating social distancing monitoring. Computer Vision, as~a sub-field of Artificial Intelligence, has been very successful in solving various complex health care problems and has shown its potential in chest CT-Scan or X-ray based COVID-19 recognition \citep{dnn2020,tougaccar2020covid} and can contribute to Social-distancing monitoring as well. Besides, deep neural networks enable us to extract complex features from the data so that we can provide a more accurate understanding of the images by analysing and classifying these features. Examples include diagnosis, clinical management and treatment, as~well as the prevention and control of COVID-19 \citep{ulhaq2020computer,nguyen2020artificial}.

Possible challenges in this area are the importance of gaining a high level of accuracy, dealing with a variety of lighting conditions, occlusion, and~real-time performance. In~this work, we aim at providing solutions to cope with the mentioned challenges, as~well.  

\noindent The main contribution of this research can be highlighted as follows:

\begin{itemize}
\item This study aims to support the reduction of the coronavirus spread and its economic costs by providing an AI-based solution to automatically monitor and detect violations of social distancing among~individuals.

\item We develop a robust deep neural network (DNN) model for people detection, tracking, and~distance estimation called {DeepSOCIAL} (Sections \ref{people}--\ref{distance}). In~comparison with some recent works in this area, such as \citep{punn2020monitoring}, we offer faster and more accurate~results. 

\item We perform a live and dynamic risk assessment, by~statistical analysis of spatio-temporal data from the people movements at the scene (Section \ref{zone}). This will enable us to track the moving trajectory of people and their behaviours, to~analyse the ratio of the social distancing violations to the total number of people in the scene, and~to detect high-risk zones for short- and long-term~periods. 

\item We back up the validity of our experimental results by performing extensive tests and assessments in a diversity of indoor and outdoor datasets which outperform the state-of-the-arts (Table~\ref{initial_models_comparison}, Figure~\ref{detection_power}). 

\item The developed model can perform as a generic human detection and tracker system, not limited to social-distancing monitoring, and~it can be applied for various real-world applications such as pedestrian detection in autonomous vehicles, human action recognition, anomaly detection, and~security~systems.  

\end{itemize}

More details and further information will be provided in the following sections. In~Section~\ref{related} we discuss about more technical related works, existing challenges, and~research gaps in the field. The~proposed methodology including the model architecture and our object detection techniques, tracking, and~red-zone prediction algorithm will be proposed in Section~\ref{methodology}. In~Section~\ref{experiments}, experimental results and performance of the system will be investigated against the state-of-the-art, followed by discussions and concluding remarks in Section~\ref{conclusion}.

\section{Related~Works}\label{related}
 
In this section, we provide a brief literature review on three types of research in this area: medical and clinical-related research, tracking technologies, and~AI-based research. Although~our research falls in the AI research category, due to the nature of the research questions, first, we will have a brief review on medical and technology-based research to have an in-depth understanding about the existing challenges. 
In Section~\ref{sec:AI} (AI-based Research) we gradually transit from object detection techniques to people detection, the~existing methodologies, and~research gaps for people detection using AI and computer~vision.

\subsection{Medical~Research}
Many researchers in the medical and pharmaceutical fields are aiming at treatment of COVID-19 infectious disease; however, no definite solution has yet been found. One the other hand, controlling the spread 
of the virus in public places is another issue, where the AI, computer vision, and~technology can step-in to~help.

A variety of studies with different implementation strategies \citep{ferguson2006,thu2020effect,choi2020optimal} have proven that controlling the prevalence is a contributing factor, and~social distancing is an effective way to reduce the transmission and prevent the spread of the virus in society. Several researchers such as \citep{eskin2019a,choi2020optimal} use Susceptible, Infectious, or~Recovered (SIR) model. SIR is an epidemiological modelling system to compute the theoretical number of people infected with a contagious disease in a given population, over~time. One of the oldest yet common SIR models is Kermack and McKendrick models introduced in 1927 \citep{kermak1991}.
\mbox{Eksin~et~al. \citep{eskin2019a}}, have recently introduced a modified model of SIR by including a social distancing parameter, which can be used to determine the number of infected and recovered~individuals.

Effectiveness of social distancing practices can be evaluated based on several standard approaches. One of the main criteria is based on the reproduction ratio, $R_o$, which indicates the average number of people who may be infected from an infectious person during the entire period of the infection~\citep{heffernan2005}.  Any $R_o > 1$ indicates an increasing rate of infection within the society and $R_o < 1$ indicates that every case will infect less than 1 person, hence, the~disease rate is considered to be declining in the target~population. 

Since the $R_o$ value indicates the disease outspread, it is one of the most important indicators for selecting social distancing criteria. In~the current COVID-19 pandemic, the~World Health Organisation estimated the $R_o$ rate would be in the range of 2--2.5 \citep{R0-2020}, which is significantly higher than other similar diseases such as seasonal flu with $R_o = 1.4$. 
In \citep{nguyen2020enabling}, a~clear conclusion is drawn about the importance of applying social distancing for cases with a high amount of $R_o$.

In another research based on the game theory on the classic SIR model, an~assessment of the benefits and economic costs of social distancing has been examined \citep{reluga2010}. The~results also show that in the case of $R_o < 1$, social distancing would cause unnecessary costs, while $ R_o \approx 2$ implies that social distancing measures have the highest economic benefits. In~another similar research, Kylie~et~al. \citep{kylie2020} investigated the relationship between the stringency of social distancing and the region's economic status. This study suggests although preventing the widespread outbreak of the virus is necessary, a~moderate level of social activities could be~allowed.

Prem~et~al. \citep{prem2020} use location-specific contact patterns to investigate the effect of social distancing measures on the prevalence of COVID-19 pandemic in order to remove the persistent path of disease outbreak using susceptible-exposed-infected-removed models (SEIR).

\subsection{Tracking~Technologies}
Since the onset of coronavirus pandemic, many countries have used technology-based solutions, to~inhibit the spread of the disease \citep{son2020,punn2020,punn2020next}.  For~example, some of the developed countries, such as South Korea and India, use GPS data to monitor the movements of infected or suspected individuals to find any possible exposure among the healthy~people. 

The India government uses the Aarogya Setu 
program to find the presence of COVID-19 patients in the adjacent region, with~the help of GPS and Bluetooth. This may also help other people to maintain a safe distance from the infected person \citep{ashok2020}. Some law enforcement agencies use drones and surveillance cameras to detect large-scale rallies and have carried out regulatory measures to disperse the population \citep{rob2017,megapix2019}. 

Other researchers such as Xin~et~al. \citep{xin2018freesense} perform human detection using wireless signals by identifying phase differences and change detection in amplitude wave-forms. However, this requires multiple receiving antennas and can not be easily integrated in all public~places. 

\subsection{AI-Based~Research}
\label{sec:AI}
The utilisation of Artificial Intelligence, Computer Vision, and~Machine Learning, can help to discover the correlation of high-level features. For~example, it may enable us to understand and predict pedestrian behaviours in traffic scenes, sports activities, medical imaging, or~anomaly detection, by~analysing spatio-temporal visual information and statistical data analysis of the images sequences~\citep{shi2020review,nguyen2020artificial}.

Among AI-Health related works, some researchers have tried to predict the sickness trend of specific areas \citep{hossain2020FluSense}, to~develop crowd counting and density estimation methodologies in public places~\citep{sindagi2018survey}, or~to determine the distance of individuals from the popular swarms \citep{brighente2019ml} using a combination of visual and geo-location cellular information. However, such research works suffer from challenges such as skilled labour or the cost of designing and implementing the~infrastructures.

On the other hand, recent advances in Computer Vision, Deep Learning, and~pattern recognition, as~the sub-categories of the AI, enable the computers to understand and interpret the visual data from digital images or videos. It also allows computers to identify and classify different types of \mbox{objects~\citep{liu2020deep,rezaei2010toward,sabzevari2008object}}. Such capabilities can play an important role in empowering, encouraging, and~performing social distancing surveillance and measurements as well. For~example, Computer Vision could turn CCTV cameras in the current infrastructure capacity into ``smart'' cameras that not only monitor people but can also determine whether people follow the social distancing guidelines or not. Such systems require very precise human detection~algorithms. 

People detection in image sequences is one of the most important sub-branches in the field of object detection and computer vision. Although~many research works have been done in human detection \citep{nguyen2016human} and human action recognition \citep{serpush2020complex}, 
the 
majority of them are either limited to indoor applications or suffer from accuracy issues under outdoor challenging lighting conditions. A~range of other research works rely on manual tuning methodologies to identify people activities, however, limited functionality has always been an issue \citep{gawande2020pedestrian}. 

Convolutional Neural Networks (CNNs) have played a very important role in feature extraction and complex object classification, including human detection. With~the development of faster CPUs, GPUs, and~extended memory capacities, CNNs allow the researchers to make accurate and fast detectors compared to conventional models. However, the~long time training, detection speed and achieving better accuracy, are still remaining challenges to be solved. Narinder~et~al. \citep{punn2020monitoring} used a deep neural network (DNN) based detector, along with Deepsort \citep{wojke2017simple} algorithm as an object tracker for people detection to assess the distance violation index---the ratio of number of people who violated the social distancing measure to the total number of the assessed group. However, no statistical analysis of the outcome of their results is provided. Furthermore, no discussion about the validity of the distance measurements is~provided.

In another study by Khandelwal~et~al. \citep{kh2020using}, the~authors have addressed the people distancing in a given manufactory. They have used MobileNet V2 network \citep{s2018mobilenetv2} as a lightweight detector to reduce computational costs, which in turn provides less accuracy comparing to some other common models. Furthermore, the~method only focuses on an indoor manufactory-setup distance measurement and does not provide any statistical assessment on the virus spread. 
Similar to the other research, no~statistical analysis is performed on the results of the distance measurement in \citep{yang2020visionbased}. The~authors have made a comparison between two common types of DNN models (You Only Look Once---YOLO and Faster RCNN 
). However, the~system accuracy has been only estimated based on a shallow comparison on different datasets with non-comparable ground~truths.

\begin{figure}[t!]
\centering
\vspace{-4mm}
\includegraphics[width = 0.9\linewidth]{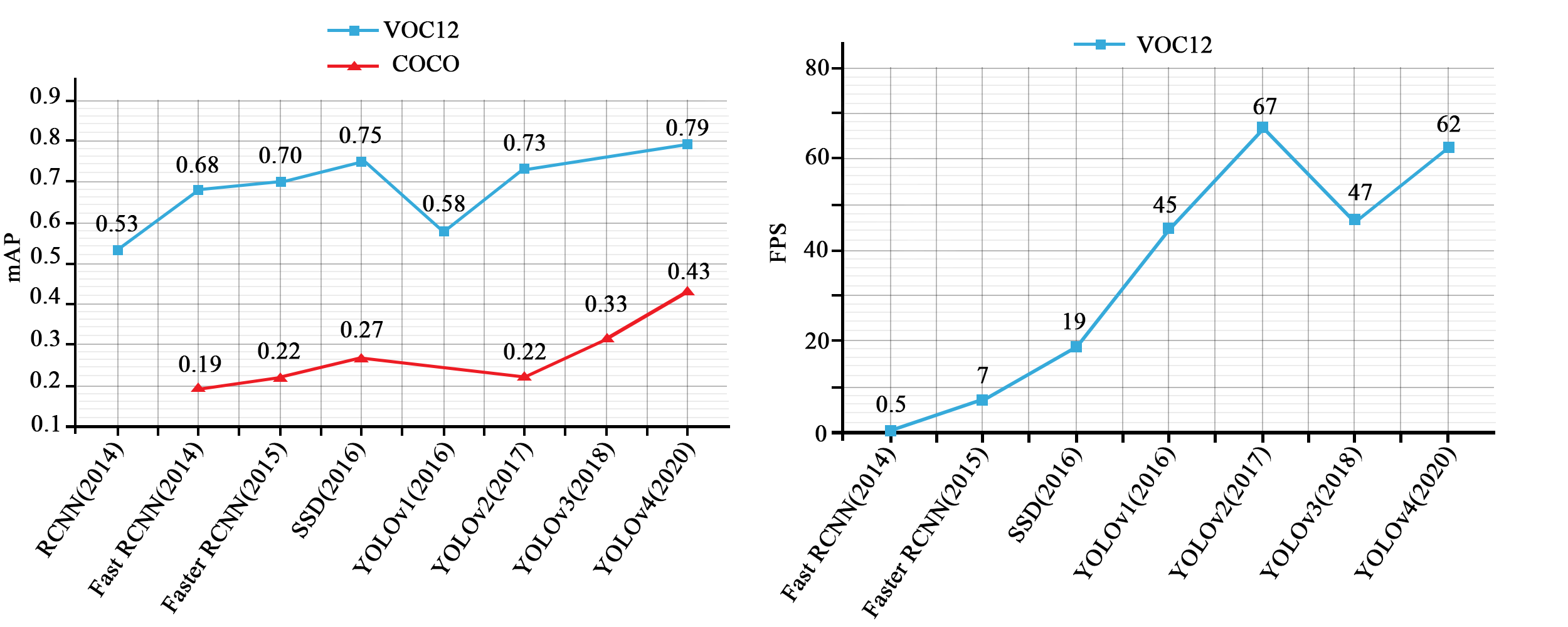}
\caption{Mean Average Precision (mAP) and Speed 
 (FPS) overview of eight most popular object detection models on Microsoft Common Objects in Context (MS-COCO) and PASCAL Visual Object Classes (VOC)~datasets.}
\label{comparison}
\end{figure}

Figure~\ref{comparison}, shows the outcome of our investigations and reviews in terms of mean Average Precisions (mAP), and~the speed (Frame Per Second---FPS) on some of the most successful object detection models such as RCNN \citep{girshick2013rich}, fast RCNN \citep{girshick2015}, faster RCNN \citep{ren2015faster}, SSD: Single Shot MultiBox Detector 
\citep{Liu_2016}, YOLOv1-v4 \citep{redmon2015look, redmon2016yolo9000, redmon2018yolov3, bochkovskiy2020yolov4} tested on the Microsoft Common Objects in Context (MS COCO)~\citep{chen2015microsoft} and PASCAL Visual Object Classes (VOC) \citep{pascal-voc-2010} data sets under similar conditions. Otherwise, the~performance of the systems may vary depending on various factors such as backbone architecture, input image size, resolution, model depth, software, and~hardware~platform.  

As can be seen from Figure~\ref{comparison}, some of the models such as SSD and YOLOv2 perform in a contradictory manner in dealing with COCO and VOC12 datasets. They may seem good in one, and~weak in another one. One of the possible reasons could be attributed to the different number of object categories in COCO and VOC12 (80 categories vs. 20).  This makes the VOC12 dataset an easier goal to learn and less challenging. However, when it comes to the higher number of classes, the~performance of the system may seem irregular, depending on the feature complexity of each object (good in some detections and weak in some other). 

Since the social distancing topic is very recent, there has not been much dedicated research regarding the accuracy of people detection and inter-people distance estimation in the crowd, no~experiment on challenging datasets has been performed, no standard comparison has been conducted on common datasets, and~no analytical studies or post-processing have been considered after the people detection-phase to analyse the risk of infection~distribution.

Considering the above-mentioned research gaps, we propose a new model which not only performs more accurate and faster than the state-of-the-art but also will be trained and tested using a large and comprehensive dataset, in~challenging environments and lighting conditions. This will ensure the model is capable of performing in real-world scenarios, particularly in covered shopping centres where the lighting conditions are not as ideal as the outdoor lighting. Furthermore, we offer post-detection and post-processing analytical solutions to mitigate the spread of the~virus.

\section{Methodology}\label{methodology}

We propose a three-stage model including people detection, tracking, inter-distance estimation as a total solution for social distancing monitoring and zone-based infection risk analysis. The~system can be integrated and applied on all types of CCTV surveillance cameras with any resolution from VGA 
to Full-HD, with~real-time~performance.

\subsection{People~Detection} \label{people}

\begin{figure}[t!]
\centering
\includegraphics[width = 1\linewidth]{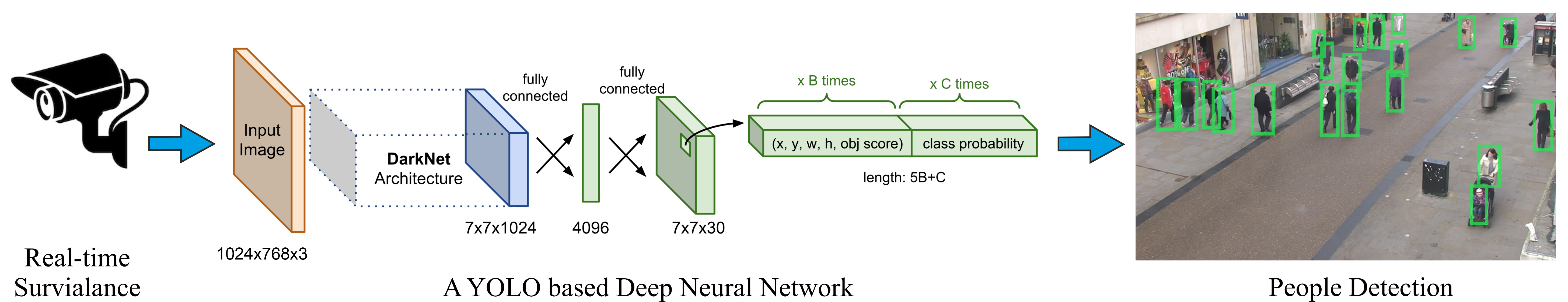}
\caption{Stage 1---The overall structure of the people detection~module.}
\label{fig-phase1}
\end{figure}

Figure~\ref{fig-phase1} shows the overall structure of the Stage 1. A~CCTV Camera collects the input video sequences, and~passes them to our Deep Neural Network model. The~output of the model would be the detected people in the scene with their unique localisation bounding boxes. 
The objective is to develop a robust human (people) detection model, capable of dealing with various types of challenges such as variations in clothes, postures, at~far and close distances, with/without occlusion, and~under different lighting~conditions. 

Modern DNN-based object detectors (such as those listed in Figure~\ref{comparison}) 
consist of three sections: {input module} and related operations such as augmentation, a~{backbone} for extracting features and a {head} for predicting classes and location of objects in the~output. 

In Table~\ref{model-options} we have summarised a comprehensive list model design options including input augmentations, state-of-the-art core object detection modules (i.e., activation functions, backbone feature extractors, neck, and~head). The~table, offers a variety of possible choices for neck, head, and~other sub-modules (depending on the requirements of the model). However, we mainly focus on the requirement of this~research. 

\begin{table}[H]
\centering
\caption{State-of-the-art options and techniques to design a Convolutional Neural Network (CNN) based model. From~left to right: input towards the~output.}
\resizebox{\textwidth}{!}{
\begin{tabular}{l|llll|ll} 
\toprule
\multicolumn{1}{c|}{\multirow{1}{*}{{\textbf{Input}}}} &
\multicolumn{4}{c|}{\multirow{1}{*}{\textbf{Detection Core} $\rightarrow$}} & 
\multicolumn{2}{c}{\multirow{1}{*}{\textbf{Head} $\rightarrow$ \textbf{Output}}} \\

\midrule
 
\multirow{1}{*}{\textbf{Augmentation}}  & 
\multirow{1}{*}{\textbf{Activation}} &
\multirow{1}{*}{\textbf{Backbone}}    & 
\multirow{1}{*}{\textbf{Neck}}  &  \multirow{1}{*}{\textbf{Regularisation}} &
\multirow{1}{*}{\textbf{Dense}}   & 
\multirow{1}{*}{\textbf{Sparse}}  \\
 \midrule

\rule{0pt}{3ex}CutOut  \citep{devries2017improved}  &  
ReLU \citep{nair2010rectified} &
VGG \citep{simonyan2014very}   & SPP \citep{He_2014} & L1, L2 \citep{ng2004feature} & 
YOLO  \citep{redmon2015look} $^*$ & R-CNN \citep{girshick2013rich} $^*$\\

MixUp \citep{zhang2017mixup}  & 
Leaky-ReLU \citep{maas2013rectifier} & 
ResNet  \citep{he2016deep}  &  
ASPP \citep{chen2017deeplab} & DropOut (DO) \citep{srivastava2014dropout} & 
SSD \citep{Liu_2016} $^*$ &   Fast-RCNN \citep{girshick2015} $^*$\\

CutMix  \citep{yun2019cutmix} & 
Param-ReLU \citep{he2015delving} &
SpineNet \citep{du2020spinenet} & 
PAN \citep{liu2018path}  &  DropPath \citep{larsson2016fractalnet} & 
RetinaNet \citep{lin2017focal} $^*$ &  Faster-RCNN  \citep{ren2015faster} $^*$ \\

Mosaic  \citep{bochkovskiy2020yolov4} & 
ReLU6 \citep{howard2017mobilenets} &
CSPResNeXt50  \citep{wang2019cspnet}  & FPN  \citep{lin2017feature}  &  Spatial DO \citep{tompson2015efficient} & 
RPN \citep{ren2015faster} $^*$ & Libra R-CNN \citep{pang2019libra}  $^*$\\

& SELU \citep{klambauer2017self} &
CSPDarknet53 \citep{wang2019cspnet}   &  
BiFPN \citep{tan2020efficientdet} & DropBlock \citep{bochkovskiy2020yolov4}  &  
CornerNet \citep{law2018cornernet} $^+$ &  Mask R-CNN \citep{he2017mask} $^*$\\

 &  Swish \citep{ramachandran2017searching} & 
 EfficientNet \citep{tan2019efficientnet} &  
 ASFF \citep{liu2019learning} &  & 
 MatrixNet \citep{rashwan2019matrix} $^+$  & CenterNet \citep{duan2019centernet} $^+$\\

 & Mish \citep{misra2019mish} &
 Darknet53 \citep{redmon2018yolov3} & 
 RFB \citep{liu2018receptive}& & 
 R-FCN \citep{dai2016r} $^*$& RepPoints  \citep{yang2019reppoints} $^+$\\
 
 &  & Inception \citep{szegedy2016rethinking,szegedy2016inception} &  SFAM \citep{zhao2019m2det} & & FCOS  \citep{tian2019fcos} $^+$ &  \\

&  & &  NAS-FPN \citep{ghiasi2019fpn}&  &  &\\
\bottomrule

\end{tabular}}\\
\begin{tabular}{ccccccc}
\multicolumn{7}{c}{\footnotesize $^*$ \footnotesize{Anchor-based}; \footnotesize  $^+$ \footnotesize{Anchor-free}.}
\end{tabular}
%
\label{model-options}
\end{table}
\vspace{-18pt}

\subsubsection{Inputs and Training~Datasets}
In order to have a robust detector, we would require a set of rich training datasets. This should include people with a variety in gender and age (man, women, boy, girl) with millions of accurate annotation and labelling. We selected two large datasets of MS COCO and Google Open Image dataset that satisfy the above-mentioed expectations, by~providing more than 3.7 million annotated people. Further, details will be provided in Section~\ref{model} (Model Training and Experimental Results). 

In YOLOv4, the~authors have dealt with two categories of training options for different parts of the network:  {``Bag of Freebies"}, which includes a set of methods to alter the model's training strategy with the aim of increasing the generalisation; and {``Bag of Specials''} which includes a set of modules that can significantly improve the object detection accuracy in exchange for a small increase in training~costs.

Among various techniques of Bag of Freebies, we used the Mosaic data augmentations \citep{bochkovskiy2020yolov4} which integrates four images into one, to~increase the size of the input data without requiring to increase the batch~size. 

On the other hand, in~batch normalisation, the~batch size reduction causes noisy estimation of mean and variance. To~address this issue, we considered the normalised values of the previous $k$ iterations instead of a single mini-batch. This is similar to Cross-Iteration Batch Normalisation (CBM)~\citep{yao2020crossiteration}.

A set of possible activation functions for BoF 
are listed in Table~\ref{model-options}. We also investigated the performance of our model against ReLU, Leaky ReLU, SELU, Swish, Parametric RELU, and~Mish. 
Our~preliminary evaluations confirmed the same results provided by Misra \citep{misra2019mish} for our human detection application. The~Mish (Equation (\ref{mish})) activation function converged towards the minimum loss, faster than Swish and ReLU, with~higher accuracy. The~result was consistent especially for diversity of parameter initialisers, regularisation methods, and~lower learning rate values.
\noindent Mish:
\\
\begin{equation}\label{mish}
\begin{aligned}
f(x) = x.\mbox{tanh}(\mbox{softplus}(x)) \\
= x.\mbox{tanh}(ln(1+e^x))
\end{aligned}
\end{equation}

\noindent with derivations:
\begin{equation}\label{dmish}
f'(x) = \frac{e^x \omega}{\delta^2}\\
\end{equation}

\noindent as a self regularised non-monotonic activation function, where
\begin{equation}
\omega = 4(x+1) + 4e^{2x} + e^{3x} + (4x + 6)  
\end{equation}

\noindent and
\begin{equation}
\delta = 2e^x + e^{2x} + 2.
\end{equation}

\subsubsection{Backbone~Architecture}
As illustrated in Figure~\ref{comparison}, YOLOv4 offers the best trade-off for the speed and the accuracy
for a multi-class object detection purpose; however, 
since YOLOv4 is an aggregation of various techniques, we undertook an in-depth study of each sub-techniques to achieve the best results for our single-class people detection model, and~to outperform the~state-of-the-art.

A basic way to improve the accuracy of CNN-based detectors is to expand the receptive field and enhance the complexity of the model using more layers; however, using this technique makes it harder to train the model. We suggest using a skip-connections technique for the ease of training, instead.

Various models use a similar policy to make connections between layers, such as Cross- Stage-Partial (CSP) connections \citep{wang2019cspnet} or Dense Blocks (consisting of Batch Normalisation, ReLU, Convolution, etc.) in DenseNet \citep{huang2016densely}. Such models have also been used in the design of some recent backbone architectures, such as  CSPResNeXt50, CSPDarknet53 \citep{wang2019cspnet} and EfficientNet-B3, which are our supported architectures options for~YOLOv4. 

Table~\ref{threemodel} summarises the brief report of our investigations for the above-mentioned backbone architectures in terms of the number of parameters and the processing speed (in \textit{fps}) for the same input size of $512 \times 512$.

Based on theoretical justifications in \citep{bochkovskiy2020yolov4} and several experiments by us, we concluded that CSPDarknet53 is the most optimal backbone model for our application, in~spite of higher complexity (due to more number of parameters). Here, the~higher number of parameters leads to the increased capability of the model in detecting multiple objects while at the same time we can maintain real-time~performance.

\subsubsection{Neck~Module}
Recently, some of the modern proposed models have placed some extra layers between the backbone and the head, called the {neck}, which is considered for feature collection from different stages of the backbone~network.

The neck section consists of several top-down and bottom-up paths to collect and combine parameters of the network in different layers, in~order to provide a more accurate image features for the head~section. 

Many CNN-based models use fully-connected layers for the classification part, and~consequently, they can only accept fixed dimensions of images as input. This can lead to two types of issues: firstly, we cannot deal with low-resolution images and secondly, the~detection of the small objects would be difficult. 
These are in contradiction with our objectives where we aim to have our model applicable in any surveillance cameras with any input image sizes and resolutions. In~order to deal with the first issue we can refer to existing methodologies such as Fully Convolutional Networks (FCNs). Such models, including YOLO (in their recent versions) have no FC-layers and therefore can deal with images with different sizes. 
However, to~cope with the second issue (i.e., dealing with small objects), we performed a pyramid technique to enhance the receptive field and extract different scales of the image from the backbone, and~finally, performing a multi-scale detection in the head~section. 

In DNNs, the~bottom layers (i.e., the first few layers) extract localised pattern and texture information to gradual build up of the semantic information which is required in the top layers. However, during~the feature extraction process, some parts of the local information that may be required for fine-tuning of the model, may be lost. In~PANet \citep{liu2018path} 
approach, the~information flow of the bottom layers will be added to top layers to strengthen of localised information; therefore, better fine-tuning and prediction can be expected. In~a recent research by Bochkovskiy~et~al. \citep{bochkovskiy2020yolov4}, it is shown that the concatenation operator performs better than addition operator to maintain the localised information and transferring  them to the top~layers.

In order to further enhancement of the receptive fields and achieve better detection power on the small objects, we consider
YOLO Feature Pyramid Network module (FPN)~\citep{lin2017feature} module for multi-scale detections. The~module extracts features in different scales from the backbone. Ref.~\citep{huang2020dc} improved YOLOv3 with a Spatial Pyramid Pooling layer (SPP) \citep{He_2014} module instead of FPN, that leads to a 2.7\% increase in the $\mbox{AP}_{50}$ on the MS COCO object detection.
The improved SPP uses max-pooling operation instead of ``Bag of Words" operation to address the issue of spatial dimensions and to deal with multi-scale detection in the head section. The~method applies a $k \times k$ max-pooling kernel, \mbox{where $k = \{1, 5, 9, 13\}$}, and~stride equals to~1.

\begin{table}[t!]
\centering
\caption{Comparisons of three backbone models in terms of number of parameters and speed (\textit{fps}) using an RTX 2070~GPU.}
\begin{tabular}{cccc} 
\toprule
\multirow{2}{*}{\textbf{Backbone Model}}  & \multirow{1}{*}{\textbf{Input}} & \multirow{1}{*}{\textbf{Number of}} & \multirow{2}{*}{\textbf{Speed (\textit{fps})}}  \\
 &\multirow{1}{*}{\textbf{Resolution}} & \multirow{1}{*}{\textbf{Parameters}}\\ 
 \midrule 
 CSPResNeXt50    & $512\times512$ & 20.6 M & 62 \\
CSPDarknet53    & $512\times512$ & 27.6 M & 66 \\
EfficientNet-B3 & $512\times512$ & 12.0 M & 26 \\ \bottomrule
\end{tabular}
\label{threemodel}
\end{table}

\begin{figure}[t!]
\vspace{2mm}
\centering
\includegraphics[width = 0.6\linewidth]{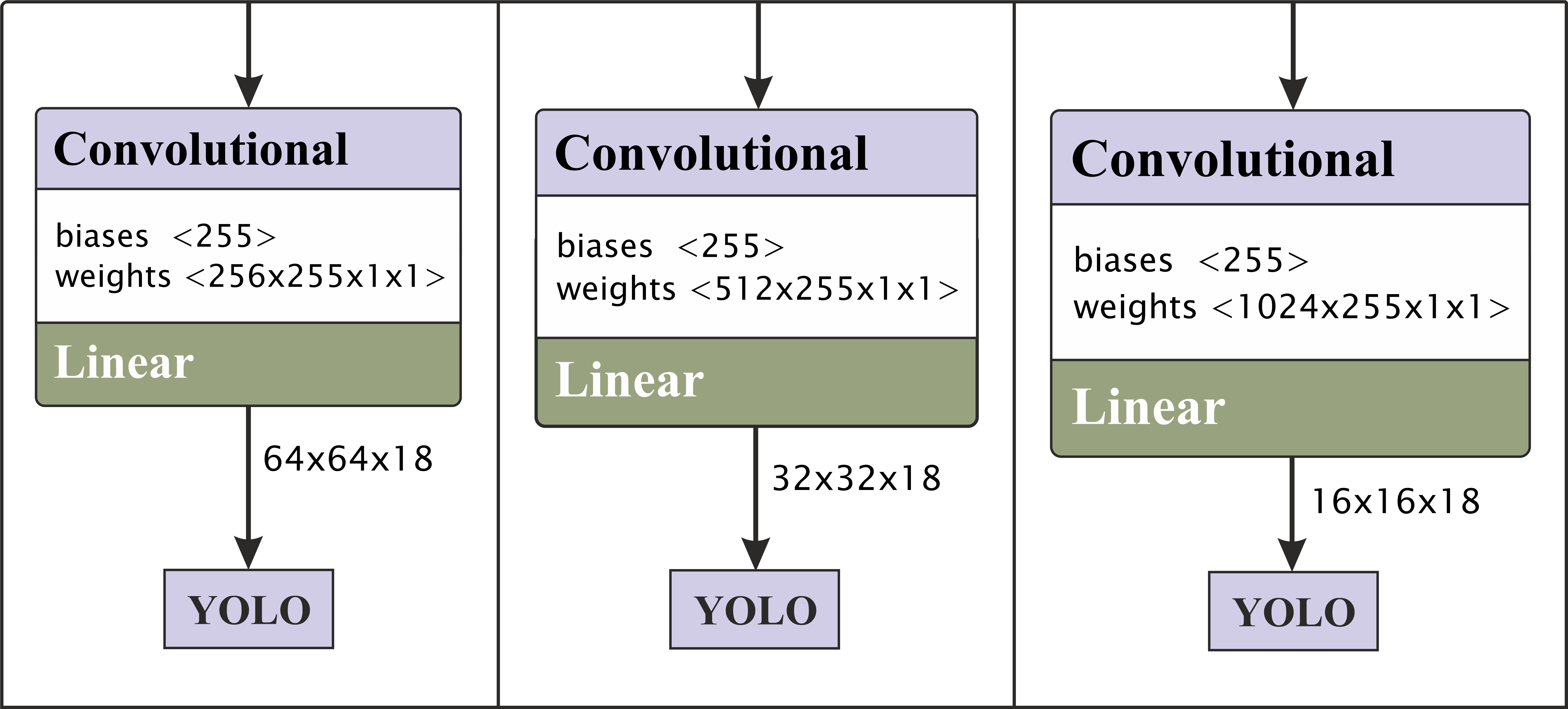}
\caption{The YOLO-based heads applied at different~scales}
\label{head}
\end{figure}

In Section~\ref{experiments} we will examine the efficiency of this approach in improving the accuracy of our YOLOv4 based model. 
Figure~\ref{head} shows the multi-scale heads that we used in our network for object detection at different~sizes.

Experimenting various rational configurations for the neck module of our model, we use spatial pyramid pooling (SPP) and PAN as well as the Spatial Attention Module (SAM) \citep{woo2018cbam} which together made one of the most effective, consistent and robust components to focus the model on optimising the~parameters.

\subsubsection{Head~Module}

The head of a DNN is responsible for classifying the objects (e.g., people, bicycles, chairs, etc.) as well as calculating the size of the objects and the coordinates of the correspondent bounding~boxes.

There are usually two types of head sections: one-stage (dense) and two-stage (sparse). The~two-stage detectors use the region proposal before applying the classification. First, the~detector extracts a set of object proposals (candidate bounding boxes) by a selective search.  Then it resizes them to a fixed size before feeding them to the CNN model. This is similar to R-CNN based detectors~\citep{girshick2013rich, girshick2015, ren2015faster}. 

In spite of the accuracy of two-stage detectors, 
such methods are not suitable for the systems with restricted computational resources  \citep{sharifi2020deephazmat}.

On the other hand, the~one-stage detectors perform a unified detection process. They map the image pixels to the enclosed grids and checks the probability of the existence of an object in each cell of the~grids. 

Similar to the work done by Liu~et~al. 
``Single Shot Multibox Detector'' (known as SSD)
, or~other works done by Redmon~et~al. \citep{redmon2015look,redmon2016yolo9000,redmon2018yolov3}, and~\mbox{Bochkovski et al. \citep{bochkovskiy2020yolov4}}, known as ``You Only Look Once'' or YOLO detectors. Such detectors use regression analysis to calculate the dimensions of bounding boxes and interpret their class probabilities.
This approach offers excellent improvements in terms of speed and~efficiency.

In the head of the model, we use the same configuration as YOLOv3. Similar to many other anchor-based models, YOLO uses predefined boxes to detect multiple objects. Then the object detection model will be trained to predict each generated anchor boxes that belong to a particular class. After~that, an~offset will be used to adjust the dimensions of the anchor box in order to better match with the ground-truth data, based on the classification and regression~loss.

Assuming the grid cell reference point $(c_x, c_y)$ at the top left corner of the object image and the bounding box prior with the width and height ($p_w, p_h$), the~network predicts a bounding box at the centre $(x\hat{}, y\hat{}\, )$ and the size of $(w\hat{}, h\hat{}\, )$ with the corresponding offset and scales of ($b_x, b_y, b_w, b_h$) as follows:
\begin{equation}
\begin{matrix}
x\hat{}  = \sigma (b_x) + c_x \\
y\hat{}  = \sigma (b_y) + c_y \\
w\hat{}  = p_w e^{b_w} \\
h\hat{}  = p_h e^{b_h}
\end{matrix}
\end{equation}

\noindent where $\sigma$ is the Sigmoid confidence score function within the range of 0 and~1. 

We represent the class of ``{human}'' with a 4-tuple $(x,y,w,h)$, where $(x,y)$ is the centre of the bounding box, and~$w, h$ are width and height, respectively.

We use three anchor boxes to find a maximum of three people in  each grid cell. Therefore, the~total number of channels is 18: (1 class + 1 object + 4 coordinates) $\times$ 3~anchors.

Since we have multiple anchor boxes for each spatial location, a~single object may be associated with more than one anchor boxes.
This problem can be resolved by using non-maximal suppression (NMS) technique and by computing intersection over union (IoU) to limit the anchor boxes~association.

As part of the weight adjustment and loss minimisation operation, we use Complete IoU (CIoU) (in Equation~(\ref{ciou})) instead of the basic IoU (Equation~(\ref{iou})). The~CIoU not only compares the location and distance of the candidate bounding boxes to the ground-truth bounding box but also it compares the aspect ratio of the size of the generated bounding boxes with the size of the ground-truth bounding~box.
\begin{equation}\label{iou}
IoU = \frac{|B \bigcap B^{gt}|}{|B \bigcup  B^{gt}|}
\end{equation}

\noindent $B^{gt} = (x^{gt}, y^{gt}, w^{gt},h^{gt})$ is the ground-truth box, and~$B = (x,y,w,h)$ is the predicted box. We use CIoU not only as a detection metric but also as a loss function:
\begin{equation}\label{ciou}
 \mathcal{L}_{CIoU} = 1 - IoU + \frac{|\rho^2(B, B^{gt})|}{c^2} + \alpha v
\end{equation}
where $\rho$ is Euclidean distance between grand-truth $B^{gt}$, and~predicted $B$ bounding box. The~diagonal length of the smallest bounding box enclosing both boxes $B$ and $B^{gt}$ is represented by $c$; and $\alpha$ is a positive trade-off parameter:
\begin{equation}
\alpha = \frac{v}{(1- IoU)+v}
\end{equation}

\noindent and $v$ measures the consistency of the aspect ratio, as~follows:
\begin{equation}
v= \frac{4}{\pi^2} \left(\arctan{\frac{w^{gt}}{h^{gt}}} - \arctan{\frac{w}{h}}\right)^2
\end{equation}

Even in case of zero-percent overlapping, the~loss functions still gives us an indication on how to adjust the weights to firstly converge the aspect size towards 1, and~secondly, how to reduce the error distance of candidate bounding boxes to the centre of the ground-truth bounding box. A~similar approach called Distance-IoU is used in \citep{zheng2019distanceiou} for another~application. 

To prevent the overfitting issue we evaluated some common regularisation techniques as shown in the Table~\ref{model-options}. Similar to the outlined results in \citep{ghiasi2018dropblock}, we found the DropBlock (DB) as one of the most effective regularisation methods in comparison with the other~options. 

In this regard, class label smoothing \citep{mller2019does} also helps to prevent over-fitting by reducing the model's confidence in the training~phase.

\begin{figure}[t!]
\centering
\vspace{2mm}
\includegraphics[width = 0.85\linewidth]{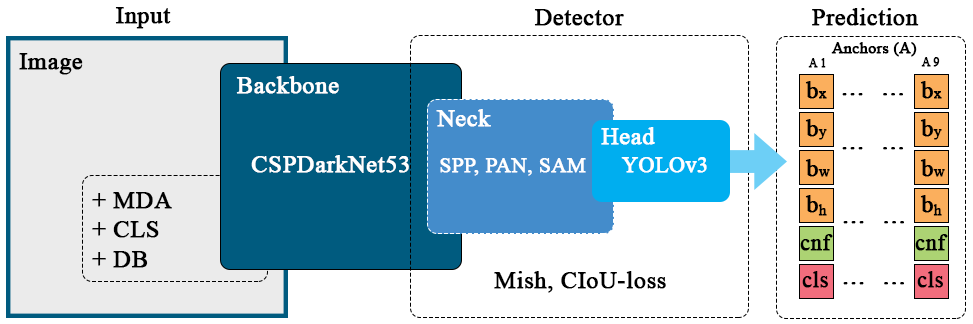}
\caption{The network structure of the proposed three-level human detection~module.}
\label{model}
\end{figure}

Figure~\ref{model} summarises the three-level structure of our human detection module in a sequence of interconnected components. In~the input part, the~Mosaic data augmentation (MDA), class label smoothing (CLS) and DropBlock (DB) regularisation are applied to the input image. In~the detector part, the~Mish activation function has been used, and~CIoU metric is considered as the loss function. In~the prediction part, for~each cell at each level, the~anchor boxes contain the information required to locate the bounding box, confidence ratio of the object, and~the corresponding class of the object. In~total, we have nine anchor~boxes.

\subsection{People~Tracking}\label{tracking}

The next step after the detection phase is people tracking and ID assignment for each~individual. 

We use the {Simple Online and Real-time} (SORT) tracking technique \citep{bewley2016simple} as a framework for the Kalman filter \citep{rezaei2017a} along with the {Hungarian} optimisation technique to track the people. Kalman filter predicts the position of the human at time $t+1$ based on the current measurement at time $t$ and the mathematical modelling of the human movement. This is an effective way to keep localising the human in case of~occlusion. 

The Hungarian algorithm is a combinatorial optimisation algorithm that helps to assign a unique ID number to identify a given object in a set of image frames, by~examining whether a person in the current frame is the same detected person in the previous frames or~not. 

\begin{figure}[t!]
\centering

\begin{tabular}{ccc}

\includegraphics[width=0.32\linewidth]{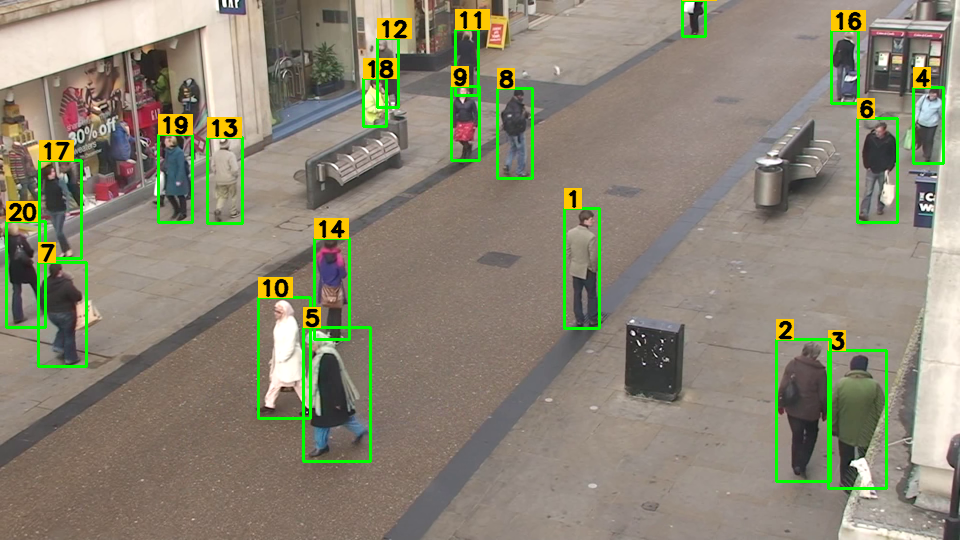} & \includegraphics[width=0.32\linewidth]{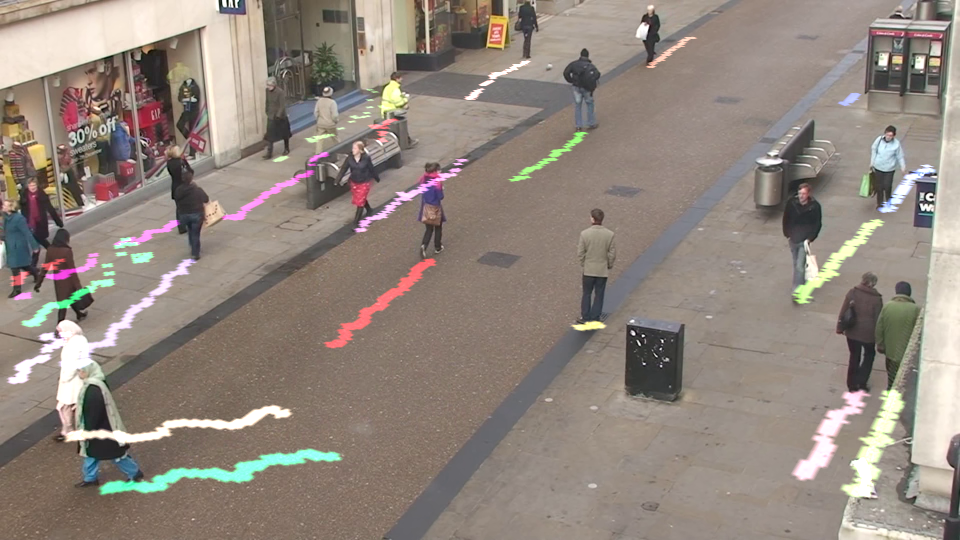} & \includegraphics[width=0.32\linewidth]{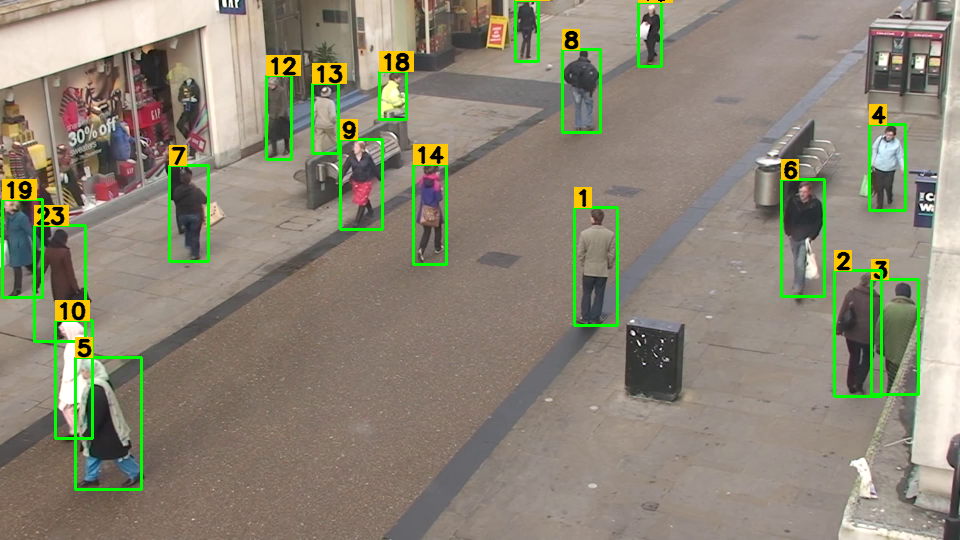} \\
(\textbf{a}) People detection and ID & (\textbf{b}) Sample tracking for  & (\textbf{c}) Final positions and IDs \\
 assignment at frame~0 & frames 0 to 100 & at frame 100 
\end{tabular}
\caption{People detection, ID assignment, tracking and moving trajectory~representation.}
\label{p-tracking}
\end{figure}

Figure~\ref{p-tracking}a shows a sample of the people detection and ID assignment, Figure~\ref{p-tracking}b illustrates the tracking path of each individual, and~Figure~\ref{p-tracking}c shows the final position and status of each individual after 100 frames of detection, tracking, and~ID assignment. We later use such temporal information for analysing the level of social distancing violations and high-risk zones of the scene.
The state of each human in a frame is modelled as:
\begin{equation}
\mbox{x} = [u, v, s, r, u', v', s']^T
\end{equation}
where $(u,v)$ represent the horizontal and vertical position of the target bounding box (i.e., the centroid); $s$ 
denotes the scale (area), and~$r$ is the aspect ratio of the bounding box sides. $u'$, $v'$, and~$s'$ are the predicted values by Kalman filter for horizontal position, vertical position, and~bounding box centroid, respectively.

When an identified human associates with a new observation, the~current bounding box will be updated with the newly observed state. This will be calculated based on the velocity and acceleration components, estimated by the Kalman filter framework. If~the predicted identities of the query individual significantly differ from the new observation, almost the same state that is predicted by the Kalman filter will be used with almost no correction. Otherwise, the~corrections weights will be split proportionally between the Kalman filter prediction and the new observation (measurement). 

As mentioned earlier, we use the Hungarian algorithm to solve the data association problem, by~calculating the IoU (Equation (\ref{iou})) and the distance (difference) of the actual input values to the predicted values by the Kalman~filter.

After the detection and tracking process, for~every input frame $I_{w \times h}$ at time $t$, we define the matrix $D_t$ that includes the location of $n$ detected human in the image carrier grid:
\begin{equation}
D_t = \{ P^t_{(x_n,y_n)} \mid x_n \in w ,\, y_n \in h \}    
\end{equation}

\subsection{Inter-Distance~Estimation} \label{distance}

Stereo-vision is a popular technique for distance estimation such as in \citep{saleem2018effects}; however, this is not a feasible approach in our research when we aim at the integration of an efficient solution, applicable in all public places using only a basic CCTV camera. Therefore we adhere to a monocular~solution. 

On the other hand, by~using a single camera, the~projection of a 3-D world scene into a 2-D perspective image plane leads to unrealistic pixel-distances between the objects. This is called perspective effect, in~which we can not perceive uniform distribution of distances in the entire image. For~example, parallel lines intersect at the horizon and farther people to the camera seem much shorter than the people who are closer to the camera coordinate~centre.

In 3-dimensional space, the~centre or the reference point of each bounding box is associated with three parameters $(x, y, z)$, while in  the image received from the camera, the~original 3D space is reduced to two-dimensions of $(x, y)$, and~the depth parameter $(z)$ is not available. In~such a lowered-dimensional space, the~direct use of the Euclidean distance criterion to measure inter-people distance estimation would be~erroneous.

In order to apply a calibrated IPM transition, we first need to have a camera calibration by setting $z = 0$ to eliminate the perspective effect. We also need to know the camera location, its height, angle of view, as~well as the optics specifications (i.e., the camera intrinsic parameters)~\citep{rezaei2017a}.

By applying the IMP, the~2D pixel points $(u, v)$ will be mapped to the corresponding world coordinate points $(X_w, Y_w, Z_w)$:
\begin{equation}\label{ipm}
[u\hspace{0.2cm} v \hspace{0.2cm} 1]^T = KRT [X_w \hspace{0.2cm} Y_w \hspace{0.2cm} Z_w  \hspace{0.2cm}1]^T
\end{equation}
\noindent where $R$ is the rotation matrix:
\begin{equation}\label{rotation}
R=
\begin{bmatrix}
1 & 0 & 0 & 0\\
0 & \cos\theta& -\sin\theta & 0\\
0 & \sin\theta & \cos\theta & 0\\
0 & 0 & 0 & 1
\end{bmatrix},
\end{equation}
\noindent $T$ is the translation matrix:
\begin{equation}\label{translation}
T=
\begin{bmatrix}
1 & 0 & 0 & 0\\
0 & 1 & 0 & 0\\
0 & 0 & 1 & -\frac{h}{\sin\theta}\\
0 & 0 & 0 & 1
\end{bmatrix},
\end{equation}

\noindent and $K$, the~intrinsic parameters of the camera are shown by the following matrix:
\begin{equation}\label{kmat}
K=
\begin{bmatrix}
f * ku & s & c_x & 0\\
0 & f * kv & c_y & 0\\
0 & 0 & 1 & 0
\end{bmatrix}
\end{equation}

\noindent where $h$ is the camera height, $f$ is focal length, and~$ku$ and $kv$ are the measured calibration coefficient values in horizontal and vertical pixel units, respectively. $(c_x, c_y)$ is the principal point shifts that corrects the optical axis of the image~plane.

The camera creates an image with a projection of three-dimensional points in the world coordinate that falls on a retina plane. Using homogeneous coordinates, the~relationship between three-dimensional points and the resulting image points of projection can be shown as follows:
\begin{equation}\label{mat1}
\\
\begin{bmatrix}
u\\
(v)\\
1
\end{bmatrix}
=
\begin{bmatrix}
m_{11} & m_{12} & m_{13} & m_{14}\\
m_{21} & m_{22} & m_{23} & m_{24}\\
m_{31} & m_{32} & m_{33} & m_{34}\\
\end{bmatrix}
\begin{bmatrix}
X_w\\
\begin{bmatrix}
Y_w\\
Z_w
\end{bmatrix}\\
1
\end{bmatrix}
\end{equation}

\noindent where $M \in \mathbb{R}^{3 \times 4}$ is the transformation matrix with $m_{ij}$  elements in Equation~(\ref{mat1}), that maps the world coordinate points into the image points based on the camera location and the reference frame, provided by the  Camera Intrinsic Matrix $K$ (Equation~(\ref{kmat})), Rotation Matrix $R$ (Equation~(\ref{rotation})) and the Translation Matrix $T$ (Equation~(\ref{translation})).

Considering the camera image plane perpendicular to the $Z$ access in the world coordinate system (i.e., $z=0$) the dimensions of the above equation can be reduced to the following form:
\begin{equation}
\begin{bmatrix}
u\\
(v)\\
1
\end{bmatrix}
=
\begin{bmatrix}
m_{11} & m_{12} & m_{13} \\
m_{21} & m_{22} & m_{23} \\
m_{31} & m_{32} & m_{33} \\
\end{bmatrix}
\begin{bmatrix}
X_w\\
(Y_w)\\
1
\end{bmatrix}
\end{equation}

\noindent and finally transferring from the perspective space to inverse perspective space (BEV) can also be expressed in the following scalar form:
\begin{equation}\label{mapmath}
\begin{aligned}
(u,v) = 
(\frac{m_{11}\times x_w+ m_{12}\times y_w + m_{13}} {m_{31}\times x_w + m_{32}\times y_w + m_{33}},  \\
\frac{m_{21}\times x_w + m_{22} \times y_w + m_{23}} {m_{31}\times x_w + m_{32} \times y_w + m_{33}})
\end{aligned}
\end{equation}

\section{Model Training and Experimental~Results}\label{experiments}

In this section we discuss the steps taken to train our human detection model and the investigated datasets to train the model, followed by experimental results on people detection, social distancing measures, and~risk infection~assessment. 

\subsection{Model~Training}

Four common multi-object annotated datasets were investigated including PASCAL VOC~\citep{pascal-voc-2010}, Microsoft COCO \citep{chen2015microsoft}, Image Net ILSVRC \citep{ILSVRC15}, and~Google Open Images Datasets V6+ \citep{kuznetsova2020} which included 16 Million ground-truth bounding boxes in 600 categories. The~dataset was a collection of 19,957~classes and the major part of the dataset was suitable for human detection and identification. The~dataset was annotated using the bounding-box labels on each image along with the corresponding coordinates of each~label.

\begin{figure}[t]
\centering
\includegraphics[width = 0.65\linewidth]{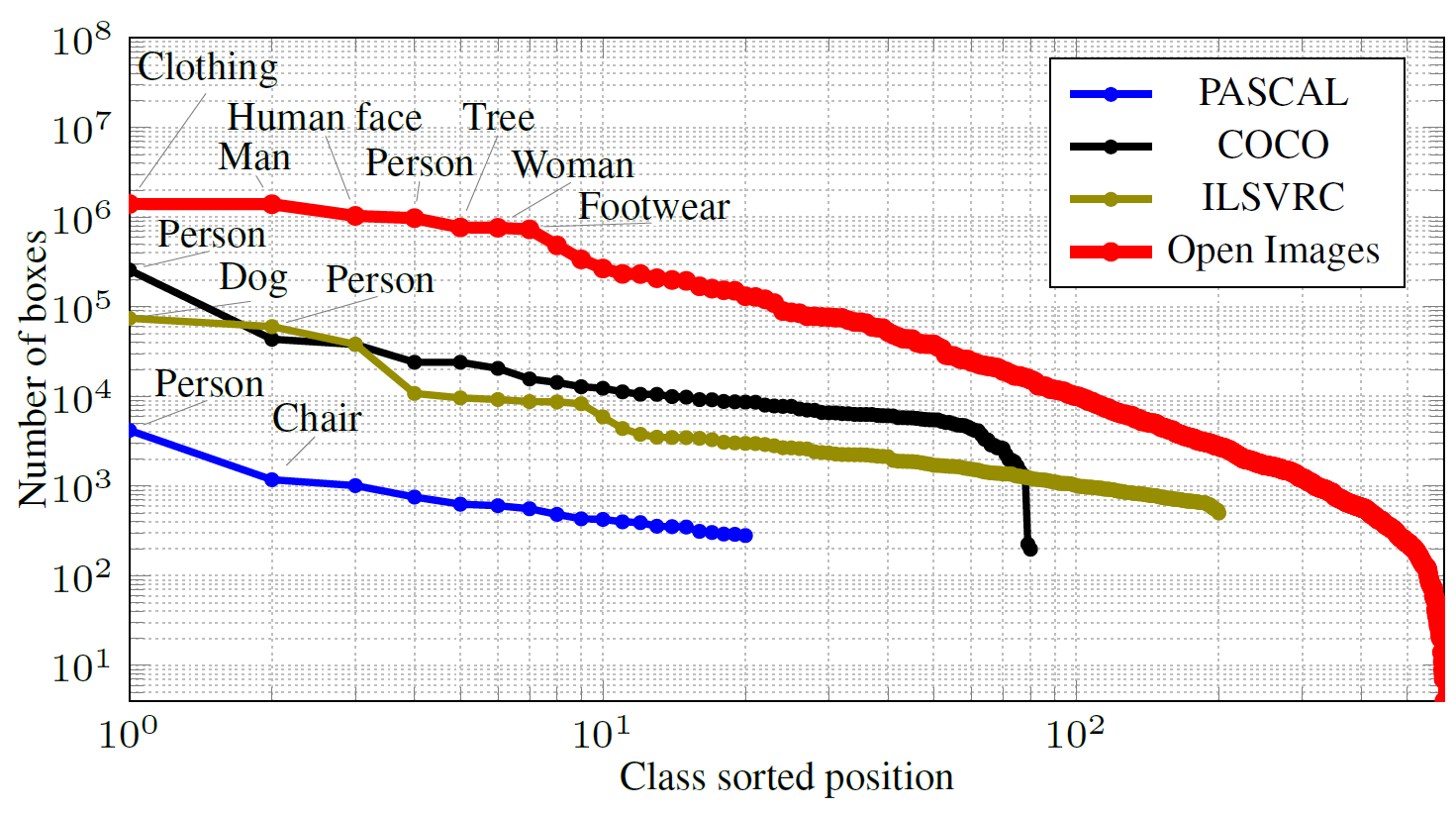}
\caption{The number of annotated boxes per class in four common datasets. The~horizontal axis is represented in logarithmic scale for better~readability.}
\label{datasets}
\end{figure}

Figure~\ref{datasets} represents the sorted rank of object classes with the number of bounding boxes for each class in each dataset. In~Google Open Images dataset (GOI) the class ``Person'' 
 shows the $4^{th}$ rank, with~nearly $10^6$ annotated bounding boxes; richer than other three investigated datasets. In~addition to the class {person}, we also adopted four more classes of {``Man'', ``Woman'', ``Boy'', and~``Girl''} from the GOI dataset for the ``human detection'' training purpose. This made a total number of 3,762,615 samples that we used from training, including 257,253 samples from the COCO dataset and 3,505,362 samples from the GOI~dataset.
 
 We also considered the category of human body parts such as the legs as we believe this allows the detector to learn a more general concept of a human being, particularly in occluded situations or in case of partial visibility e.g.,~at the borders of the input image where the full-body of the individuals can not be~perceived.

\begin{figure}[t!]
\centering
\includegraphics[width = 0.85\linewidth]{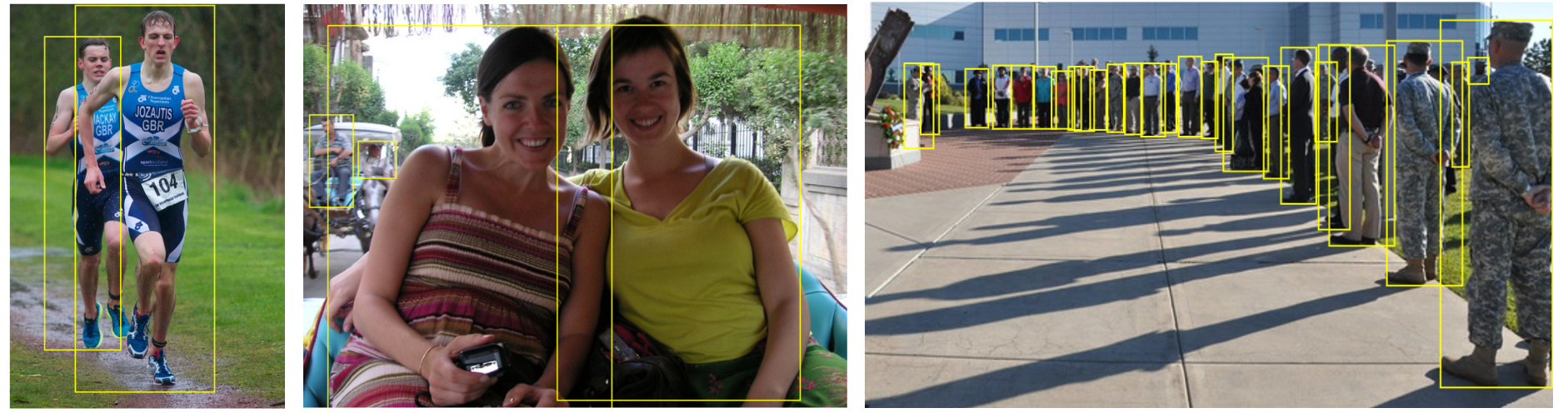}
\caption{Examples of annotated images in Open Images~dataset.}
\label{fig-GOI}
\end{figure}

Figure~\ref{fig-GOI} shows examples of annotated images from the Open Images Dataset. The~figure illustrates the diversity of the annotated people including large and small bounding boxes, in~far and near distances to the camera image plane, people occlusion, as~well as variations in shades and lighting~conditions. 

In order to train the developed model, we considered a transfer learning approach by using pre-trained models on Microsoft COCO dataset \citep{chen2015microsoft} followed by fine-tuning and optimisation of our YOLO-based~model.

We also used Stochastic Gradient Descent (SGD) with \textit{warm restarts}
~\citep{loshchilov2016sgdr} to change the learning rate during the training process. This helped to jump out of local minima in the solution space and save the training time. The~method initially considered a large value of the learning rate, then slowed down the learning speed halfway, and~eventually reduced the learning rate for each batch, with~a tiny downward slope. We decreased the learning rate using a cosine annealing function for each batch as~follows:
\begin{equation}
\eta_t^i = \eta_{min}^i  + \frac{1}{2} (\eta_{max}^i - \eta_{min}^i )(1+ \cos(\frac{T_{cur}}{T_i} \pi))
\end{equation}
where $\eta_t$ is the current learning rate in the $i$th run, $\eta_{min}$ and $\eta_{max}$ are the minimum and maximum target learning rates. $T_{cur}$ is the number of epochs executed since the last restart, and~$T_i$ is the number of epochs performed since the restart of the~SGD.

\subsection{Performance~Evaluation}

In order to test the performance of the proposed model, we used the Oxford Town Centre (OTC) dataset \citep{megapix2019} as a previously unseen and challenging dataset with very frequent cases of occlusions, overlaps, and~crowded zones. The~dataset also contained a good diversity of  human specimens in terms of clothes and appearance in a real-world public~place.

In order to provide a similar condition for performance analysis of YOLO based models, we~fine-tuned each model on human categories of the Google Open Images (GOI) \citep{kuznetsova2020} data set. This was done by removing the last layer of each model and placing a new layer (with random values of the uniform probability distribution) corresponding to a binary classification (presence or absence of a human). Furthermore, in~order to provide an equal condition for the speed and generalisability, we~also tested each of the trained models against the OTC dataset \citep{megapix2019}. 

We evaluated and compared our developed models against three common metrics of object detection in computer vision, including Precision Rate, Recall Rate, and~FPS against three state-of-the-art human/object detection methods. 

All of the benchmarking tests and comparisons were conducted on the same hardware and software: a Windows 10-based platform with an Intel$^{\copyright}$ Core\texttrademark{} i5-3570K processor and an NVIDIA RTX 2080 GPU with CUDA version~10.1.

In terms of mass deployment of the system, the~above hardware setup can handle up to 10 input cameras for real-time monitoring of e.g.,~different floors and angles of large shopping malls. However, for~smaller scales, a~cheaper RTX 1080 GPU or an 8-core/16-thread 10th generation Core\texttrademark\, i7 CPU would suffice for real-time~performance. 

\begin{figure}[t!]
\centering
\vspace{2mm}
\includegraphics[width = 0.7\linewidth]{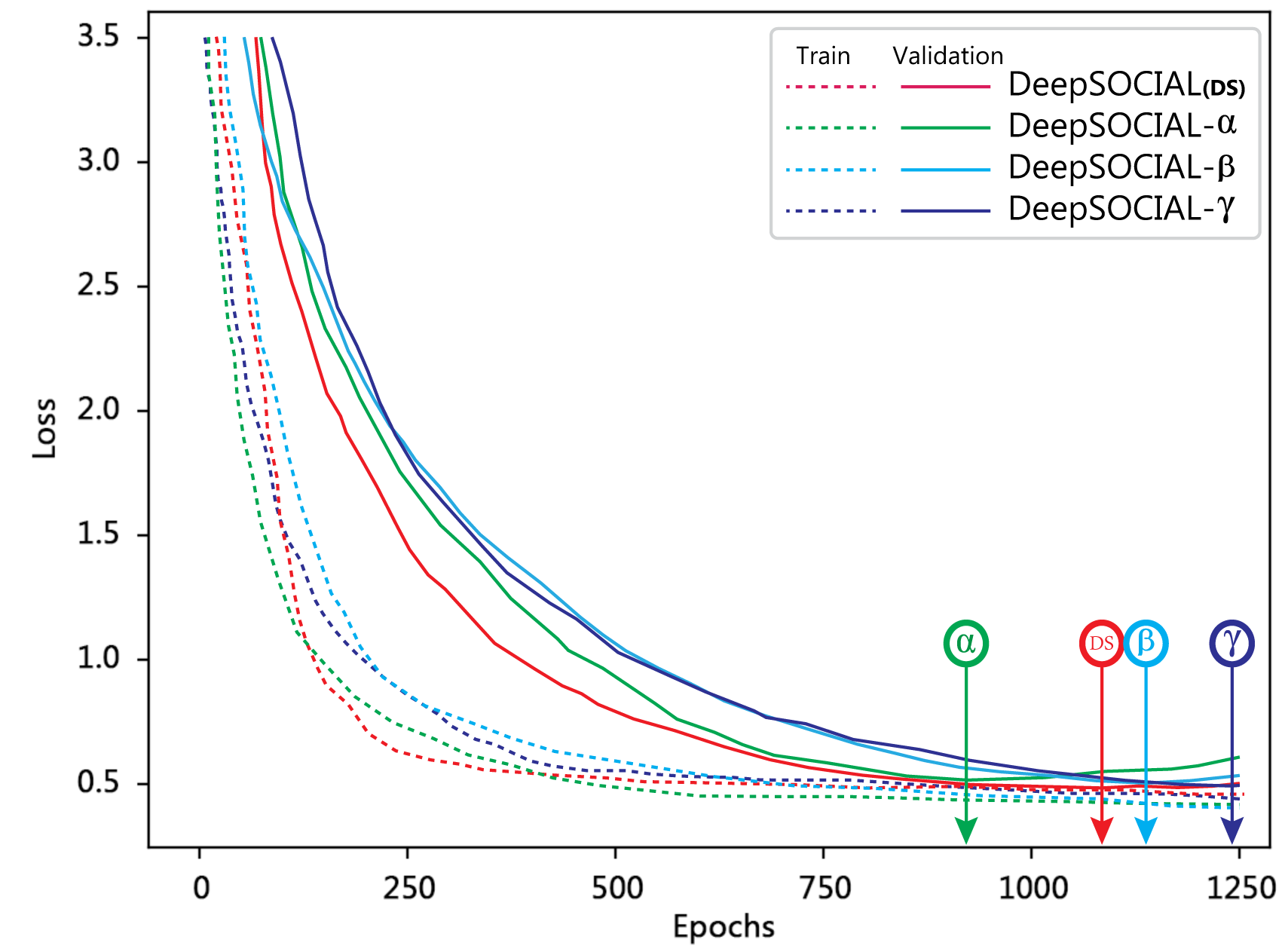}
\caption{Training and validation loss of the DeepSOCIAL models over the Open Images~dataset.}
\label{loss}
\end{figure}

Figure~\ref{loss} illustrates the development of loss function in training and validations phases for four versions of our DeepSOCIAL model with different backbone structures. The~graphs confirm a fast yet smooth and stable transition for minimising the loss function in DS version after 1090 epochs where we reached to an optimal trade-off point for both the training and validation loss. Table~\ref{initial_models_comparison} provides the details of each backbone and the outcome of the experimental results against three more state-of-the-art model on the OTC~Dataset. 

\unskip

\begin{table}[H]
\centering
\caption{Accuracy, recall-rate,
 and speed comparison for seven Deep Neural Network (DNN) models on the Oxford Town Centre~dataset.}
\renewcommand\arraystretch{1.2}
\begin{tabular}{ c c c c c c } 
\toprule
\multirow{1}{*}{\textbf{Method}}  & \multirow{1}{*}{\textbf{Backbone}} &  \multirow{1}{*}{\textbf{Precision}} & \multirow{1}{*}{\textbf{Recall}} & \multirow{1}{*}{\textbf{FPS}}\\
\midrule
DeepSOCIAL-$\alpha$ & CSPResNeXt50-PANet-SPP-SAM & \textcolor[rgb]{0,0,1}{99.8\%} & 96.7 & 23.8 \\
DeepSOCIAL-$\beta$ & CSPDarkNet53-PANet-SPP    & 99.5\% & 97.1 & \textcolor[rgb]{0,0,1}{24.1}\\
DeepSOCIAL-$\gamma$ & CSPResNeXt50-PANet-SPP    & 99.6\% & 96.7 & 23.8\\
\textbf{DeepSOCIAL (DS)} 
& CSPDarkNet53-PANet-SPP-SAM   & \textcolor[rgb]{0,0,1}{99.8\%} & \textcolor[rgb]{0,0,1}{97.6} & \textcolor[rgb]{0,0,1}{24.1}\\
YOLOv3 & DarkNet53     & 84.6\% & 68.2 & 23 \\
SSD & VGG-16      & 69.1\% & 60.5& 10  \\
Faster R-CNN & ResNet-50 & 96.9\%  & 83.0& 3\\\bottomrule
\end{tabular}
\vspace{3mm}
\label{initial_models_comparison}
\end{table}

\begin{figure}[t!]
\centering
\begin{tabular}{ccc}

\includegraphics[width=0.346\linewidth]{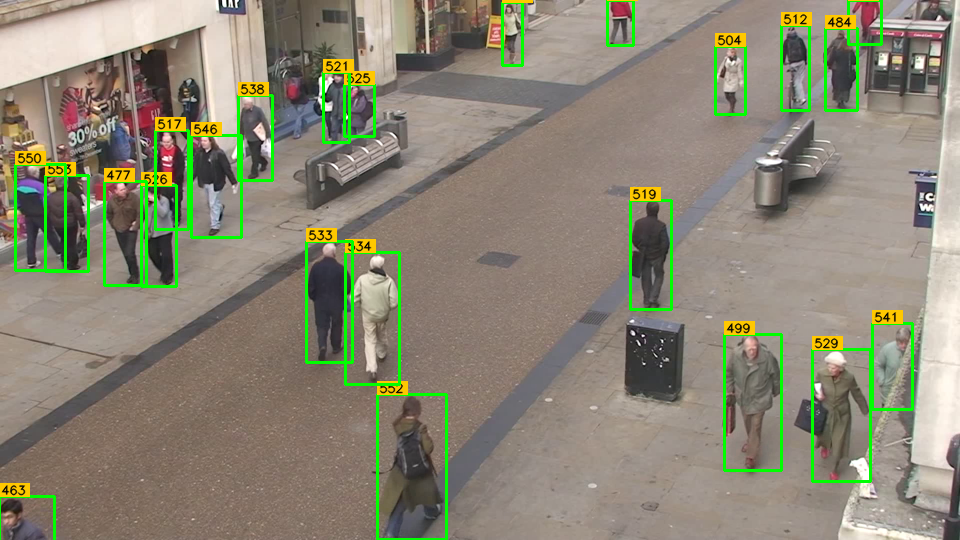} & \includegraphics[width=0.27\linewidth]{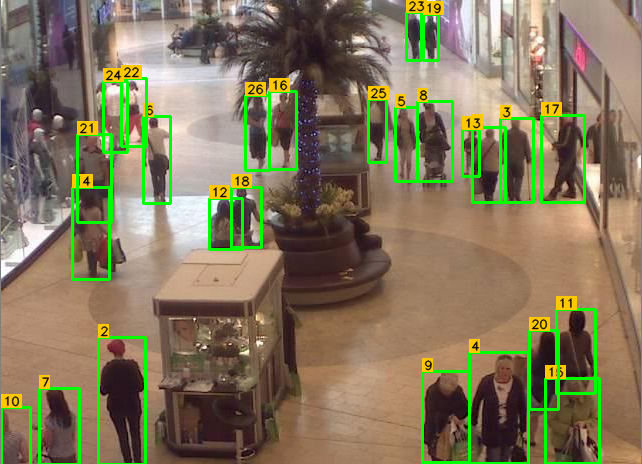} & \includegraphics[width=0.293\linewidth]{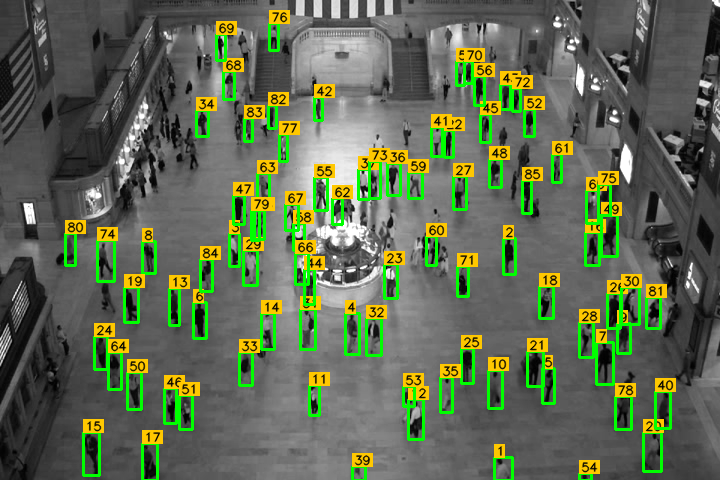} \\
(\textbf{a}) Oxford Town Centre \citep{megapix2019} & (\textbf{b}) Mall Dataset \citep{chen2012feature}  & (\textbf{c}) Train Station \citep{zhou2012understanding} \\
\end{tabular}
\caption{Detection performance of the DeepSOCIAL model in three different datasets, from~a low resolution of $640\times480$ to an HD resolution of $1920\times1080$.}
\label{detection_power}
\end{figure}

Figure~\ref{detection_power} visualises the robustness of the proposed detectors in three challenging indoor/outdoor publicly available datasets: Oxford Town Centre, Mall Dataset, and~Train Station~Dataset.

Interestingly, the~Faster-RCNN model showed good generalisability; however, its low speed was an issue which seems to be due to the computational cost of the ``region proposal'' technique. Since the system required a real-time performance, any model with the speeds slower than 10 \textit{fps} and/or a low level of accuracy may not be a suitable option for Social Distancing monitoring. Therefore, SSD and Faster-RCNN failed in this benchmarking assessment, despite their popularity in other applications. YOLOv3 and YOLOv4 based DeepSOCIAL--\textit{X} methods provided relatively better results comparing the other models, and~finally, the~proposed DeepSOCIAL-DS model outperformed all of the assessed models in terms of both speed and~accuracy. 

\begin{figure}[t!]
\centering
\includegraphics[width = 1\linewidth]{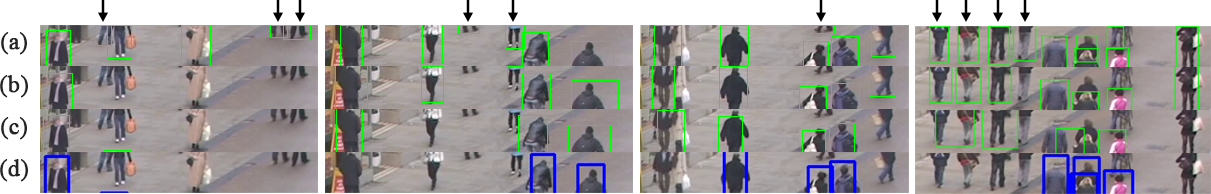}
\caption{Human detection with partial visibility (missing upper body parts). (\textbf{a}) DeepSOCIAL (\textbf{b}) YOLOv4 trained on MS-COCO, (\textbf{c}) YOLOv3 (\textbf{d}) Ground-truth annotations from the Oxford Town Centre (OTC)~dataset.}
\label{foottest}
\end{figure}

Figure~\ref{foottest} provides sample footage of the challenging scenarios when the people either entered or exited the scene, and~only part of their body (e.g., their feet) was visible. The~figure clearly indicates the strength of DeepSOCIAL in Row (a), comparing to the state-of-the-art. The~bottom row (d) with blue bounding boxes shows the ground-truth while some of the existing people with partial visibility are not annotated even in original ground-truth dataset. Row (c), YOLOv3 shows a couple of more detections; however, the~IoU of the suggested bounding boxes are low and some of them can be counted as false positives. Row (b), the~standard YOLOv4-based detector, shows a significant improvement comparing to row (c) and is considered as the second-best. Row (a), the~DeepSOCIAL, shows 10 more true positive detections (highlighted by vertical arrows) comparing to the second best~approach. 

Although the DeepSOCIAL model showed superior results even in challenging scenarios such as partial visibility and truncated objects, there could be some further challenges such as detections in extreme lighting conditions and lens distortion effects that may affect the performance of the model. This requires further investigations and experiments. Unfortunately, at~the time of this research, there were no such datasets publicly available for our~assessment.

\subsection{Social Distancing~Evaluations}
\label{SDE}

We considered the midpoint of the bottom edge of the detected bounding boxes as our reference points (i.e., shoes' location). After~the IPM, we would expect to have the location of each person, in~the homogeneous space of BEV with a linear distance~representation. 

Any two people $P_i, P_j$ with the Euclidean distance of smaller than $r$ (i.e., the set restriction) in the BEV space were considered as contributors in social distancing violation: 

Depending on the type of overlapping and the violation assessment criteria, we define a violation detection function $V$ with the input parameters of a pixel metrics $\xi$, the~set safe distance of $r$ (e.g., $6ft$ or $\approx$2 m), the~position of the query human $H_q$, and~the closest surrounding person $P_o$.
\begin{equation}\label{vioform}
V= \Lambda_{\xi}(H_q, P_o, r)
\end{equation}

\noindent where $\xi$ is the number of pixels in a line that represent a 1.0-m length in the BEV~space. 

\begin{figure}[t!]
\centering
\vspace{3mm}
\includegraphics[width=1\linewidth]{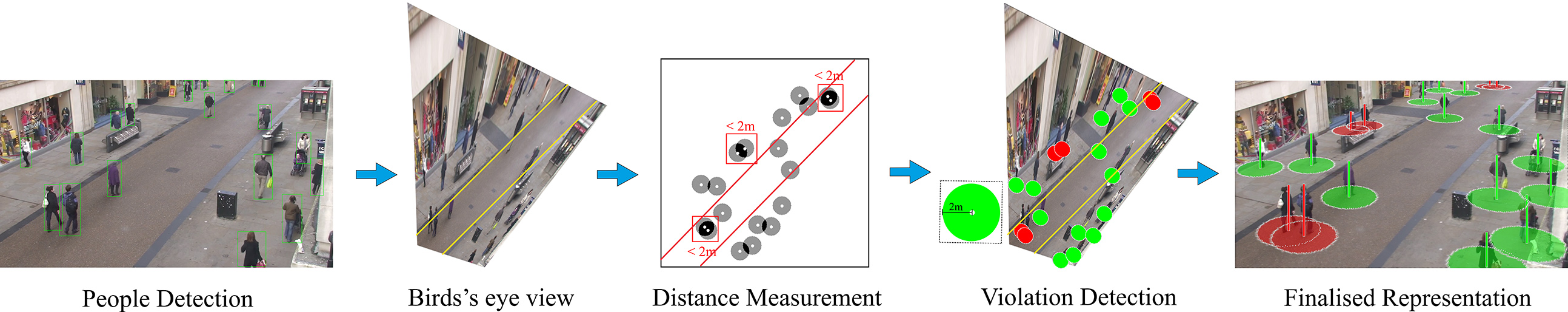}
\caption{The summary of the steps taken in Sections~\ref{people}--\ref{distance} for people detection, tracking, and~distance~estimation.}
\label{fig-bird}
\end{figure}

Figure~\ref{fig-bird} left  (from Oxford Town Centre Dataset \citep{megapix2019}) shows the detected people followed by the steps we have taken for inter-people distance estimation including tracking, IPM, homogeneous $360^\circ$ distance estimation, safe movements (people in green circles) and the violating people (with~overlapping red circles):
\begin{equation}\label{norm}
\Lambda_{\xi}(P_i, P_j, r) = \left\{\begin{matrix}
1 & \mbox{if   }  \sqrt{(x_i-x_j)^2 + (y_i-y_j)^2}  \leq r \\ 
0& \mbox{if   } \sqrt{(x_i-x_j)^2 + (y_i-y_j)^2}  > r 
\end{matrix}\right. 
\end{equation}

Regarding the Oxford Town Centre (OTC) dataset, every 10 pixels in the BEV space was equivalent to 98 cm in the real world. Therefore, $r \approx 2 \times \xi$ and equal to 20 pixels. The~inter-people distance measurement was measured based on the Euclidean $L2$ norm distance (Equation~(\ref{norm})).

One of the controversial opinions that we received from health authorities was the way of dealing with family members and couples in social distancing monitoring. Some researchers believed social distancing should apply on every single individual without any exceptions and others were advising the couples and family members can walk in a close proximity without being counted as a breach of social distancing. In~some countries such as in the UK and EU region the guideline allows two family members or a couple walk together without considering it as the breach of social distancing. We also considered a solution to activate the couple detection. This will be helpful when we aim at recognising risky zones based on the statistical analysis of overall movements and social distancing violations over a mid or long period (e.g., from few hours to few days).

\begin{figure}[t!]
\centering

\begin{tabular}{ll}

\includegraphics[width=0.47\linewidth]{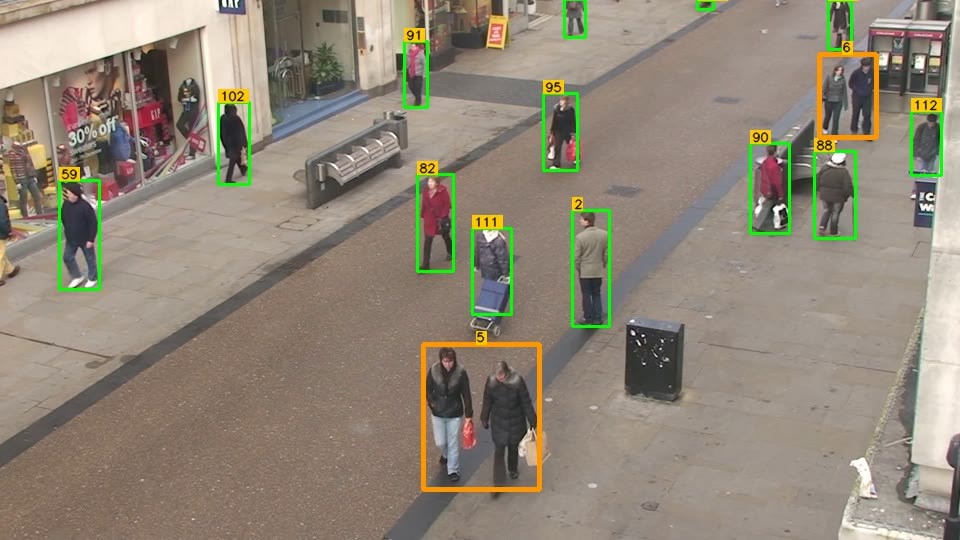} & \includegraphics[width=0.47\linewidth]{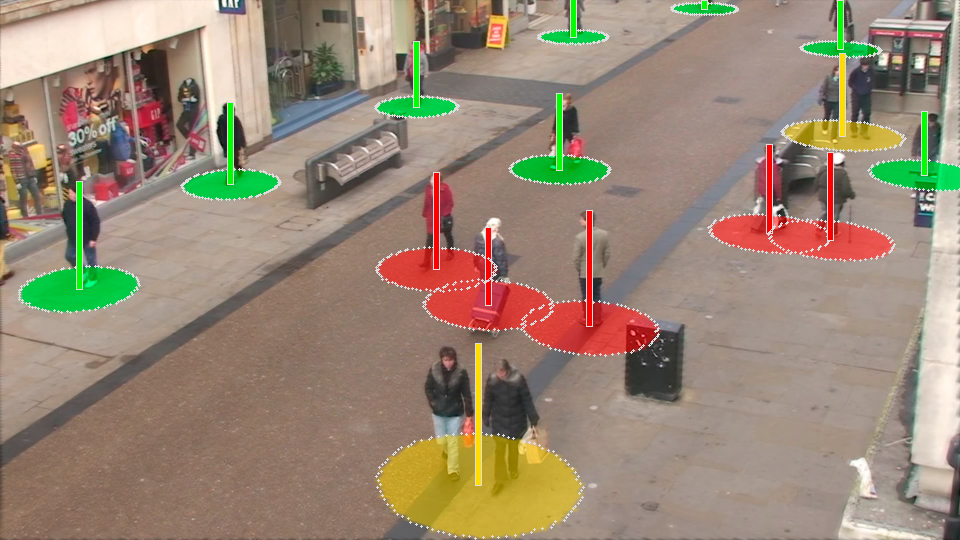}  \\
(\textbf{a}) Examples of coupled people detections- & (\textbf{b}) Three types of detections: safe, violations,   \\
 orange bounding boxes & coupled 
\end{tabular}

\caption{Social distancing violation detection for coupled people and~individuals.}
\label{couples}
\end{figure}

Applying a temporal data analysis approach, we consider two individuals $(p_i, p_j)$ as a couple, if~they are less than $d$ meters apart in an adjacency, for~a $t_{\Lambda}$ of more than $\varepsilon$ seconds.
As an example, in~Figure~\ref{couples}a, we have identified people who have been less a meter apart from each other for more than $\varepsilon=5$ s, in~the same moving trajectory:
\begin{equation}
\begin{aligned}
  C(p_i, p_j) = 1 \quad \mbox{IF} \quad p_i, p_j \in D \\
	\mbox{AND} \quad \Lambda_{\xi}{(p_i,p_j, d)} =1 \\
	\mbox{AND} \quad t_{\Lambda}(p_i,p_j) > \varepsilon 
	\end{aligned}
 \end{equation}

Figure~\ref{couples}b shows a sample of our representation for detected couples in a scene as well as multiple cases of social distancing violations. The~yellow circles drawn for people diagnosed as couples have a radios of $(2+\frac{d_c}{2})$ meter to ensure a minimum safety distance of 2 m for each of them to their left and right neighbours. $d_c$ is the distance of coupled members to each~other.

In cases where a breach occurs between  two neighbouring couple, or~between a couple and an individual, all of the involved people will turn to red status, regardless of being a couple or~not.

The flexibility of our algorithm in considering different types of scenarios enables the policymakers and health authorities to proceed with different types of investigations and evaluations for the spread of the~infection. 

\begin{figure}[t!]
\vspace{2mm}
\centering
\includegraphics[width = 1\linewidth]{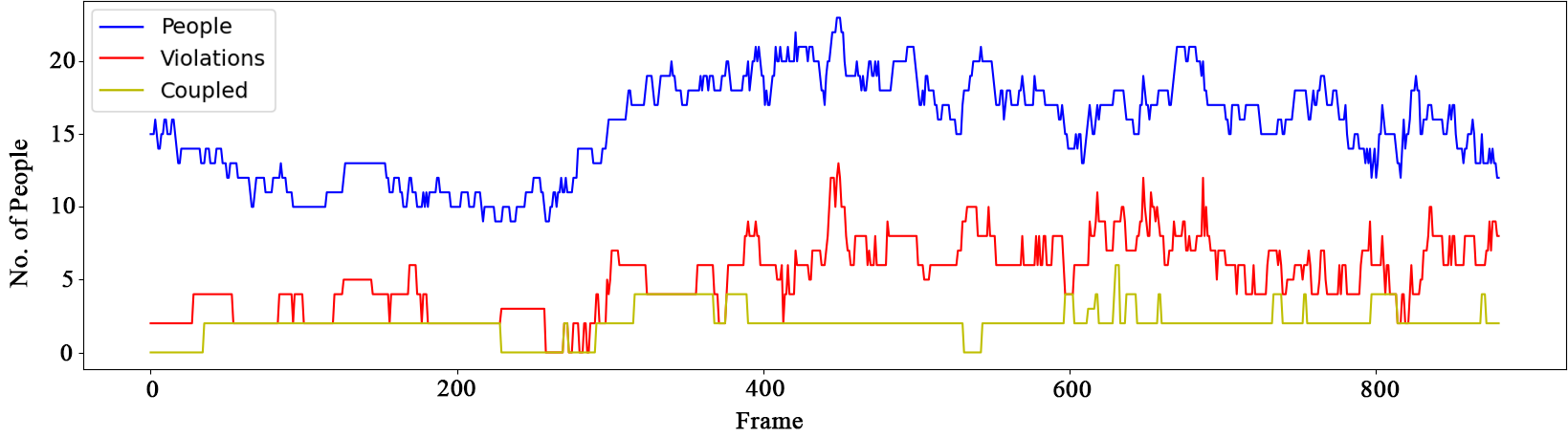}
\vspace{-5mm}
\caption{A 2D recording of the number of detected people in 900 frames from the OTC Dataset, as~well as the number of violations and number of couples with no~violations.}
\label{Statistics}
\end{figure}

For example, Figure~\ref{Statistics} from the Oxford Town Centre dataset, provides a basic statistics about the number of people in each frame, the~number of people who do not observe the distancing, the~number of social distancing violations without counting the coupled groups as~violations. 

Regarding the coupled group we reached to an accuracy and recall rate of 98.7\% and 23.9 $fps$, respectively which is slightly lower than our results of normal human detection as per the Table~\ref{initial_models_comparison}. This was expected due to added complexity in tracking two side by side people and possibly more complex occlusion~scenarios.

\subsection{Zone-Based Risk~Assessment} \label{zone}

We also tested the effectiveness of our model in assessing the long-term behaviour of the people. This can be valuable for health sector policymakers and governors to make timely decisions to save lives and reduce the consequent costs. Our experiments provided very interesting results that can be crucial to control the infection rates before it raises uncontrolled and~unexpectedly.

In addition to people inter-distance measurement, we considered a long-term spatio-temporal zone-based statistical analysis by tracking and logging the movement  trajectory of people, density of each zone, the~total number of people who violated the social-distancing measures, the~total time of the violations for each person and as the whole, identifying high-risk zones and ultimately, creating an informative risk~heat-map.

In order to perform the analysis, a~2-D grid matrix $G_t \in \mathbb{R}^{w \times h}$ (initially filled by zero) was created to keep the latest location of individuals using the input image sequences. $G_t$ represents the status of the matrix at time $t$ and $w$ and $h$ are the width and height of the input image $I$, respectively. 

To consider environmental noise and better visualisation of the intended heat-map, every person was associated with a $3 \times 3$ Gaussian kernel $k$:
\begin{equation}
k(i,j) = 
\begin{bmatrix}
0_{i-1, j-1} &  1_{i, j-1} & 0_{i+1, j-1} \\
1_{i-1, j} & 2_{i, j} & 1_{i+1, j} \\
0_{i-1, j+1} & 1_{i, j+1} &  0_{i+1, j+1}
\end{bmatrix}
\end{equation}

\noindent where $i$ and $j$ indicate the centre point of the predicted box for each individual, $p$.

The grid matrix $G$ will be updated for every new frame and accumulates the latest information of the detected people. 
\begin{equation}
G_t = G_{t-1} + P^{\,t}_{(x,y)} \quad \forall P \in D_t
\end{equation}

\begin{figure}[t!]
\centering
\begin{tabular}{ccc}

\includegraphics[width=0.28\linewidth]{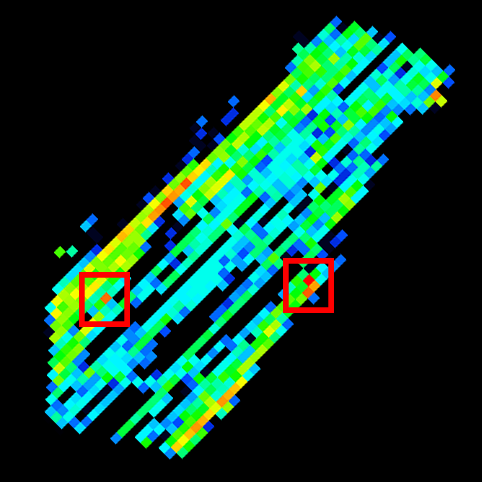} & \includegraphics[width=0.49\linewidth]{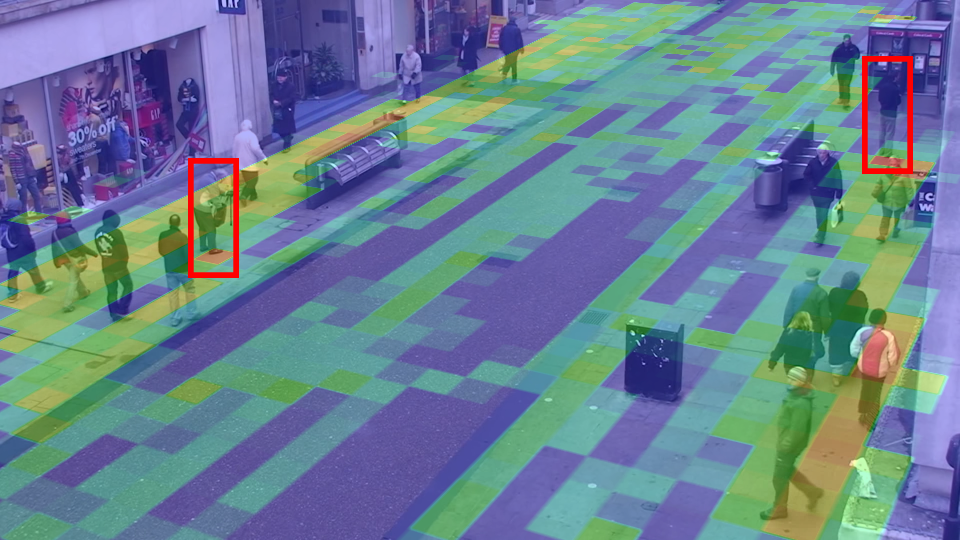}  & \includegraphics[width=0.029\linewidth]{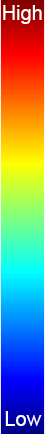}\\
(\textbf{a}) BEV heat map & (\textbf{b}) 2D Mosaic heat map  \\
\end{tabular}
\caption{Accumulated tracking maps after $500$ frames. Blue: low-risk, Red: high-risk}
\label{moveMap}
\end{figure}

Figure~\ref{moveMap} shows a sample representation of the accumulated tracking map after 500 frames of continuous people~detection. 

Since COVID-19 is an airborne virus, breathing by any static person in a fixed location can increase the density of the contamination on that point (assuming the person may carry the COVID-19), particularly in covered places with minimal ventilation. Therefore we can assign a more contamination weight to the steady-state~people. 

Figure~\ref{moveMap}a,b shows two instances of cases where two people wew steady in two particular locations of the grid for a long period; hence, the~heat map was turning red for those locations. Both side-walks also showed stronger heat map than the middle of the street due to the higher traffic of people movements. In~general, the~more red grids potentially indicate more risky~spots. 

In addition to people raw movement and tracking data, it would be more beneficial to analyse the density and the location of the people who particularly violated the social distancing~measures. 

In order to have a comprehensive set of information, we aimed to represent a long term heat-map of the environment based on the combination of accumulated detections, movements, steady-state people, and~total number of breaches of the social-distancing. This helped to identify risky zones, or~redesign the layout of the environment to make it a safer place, or~to apply more restrictions and/or limited access to particular zones. 
The restriction rules may vary depending on the application and the nature of the environment (e.g., this can vary in a hospital compared to a school). 

Applying the social distancing violation criteria as per Equation~(\ref{vioform}), we identified each individual in one of the following categories:

\begin{itemize}

\item  \textbf{Safe} 
: All people who observed the social distancing (green circles).
\begin{equation}
  Z_{g}^t = \{P^t | V_P^t = 0\}
    \end{equation}
		
\item \textbf{High-risk}: All people who violated the social distancing (red circles).
\begin{equation}
  Z_{r}^t = \{P^t | V_P^t = 1\}
    \end{equation}
		
\item  \textbf{Potentially risky}: Those people who moved together (yellow circles) and were identified as coupled. Any two people in a coupled group were considered as one identity as long as they did not breach the social distancing measures with their neighbouring people.
\begin{equation}
  Z_{y}^t = \{(P^t  | P \in C , P \not\in Z_{r}\}
    \end{equation}
		
\end{itemize}

We also used a 3-D violation matrix $S\in \mathbb{R}^{w \times h \times 3}$  to record the location of breaches for each person and its type (red or yellow):
\begin{equation}\label{heatmap-f}
\begin{aligned}
S ^t_{(x,y)} = & S_{(x,y)} ^{t-1} + \sum_{t=1}^n \left( R^{\,t}_{(x,y)}\cdot\alpha + T^{\,t}_{(x,y)}\cdot\beta + Y^{\,t}_{(x,y)}\cdot\delta \right) \\
& \forall R \in Z_r, \, T \in G_t, \, Y \in Z_y
\end{aligned}
\end{equation}

\noindent where $R$, $T$, and~$Y$ indicate cases with a violation, tracked people, and~couples, respectively. $\alpha$, $\beta$, and~$\delta$ are the relative coefficients that can be set by health-related researches depending on the importance of each factor in spreading the~virus.

In order to visualise the 3D heat map of safe and red zones, we normalised the collected data in Equation~(\ref{heatmap-f}) as follows:
\begin{equation}
N(X, l, u) = l + (u-l) \times \frac{X - min(X)}{max(X) - min (X)}
\end{equation}
where X is the non-normalised values, $l$ and $u$ are the lower bound and upper bound of the normalisation matrix that we used to define the Hue colour range on the HSV~channel. 

\begin{figure}[t!]
\centering
\subfloat[Moving trajectory and tracking map]{
\resizebox*{4.9cm}{!}{\includegraphics{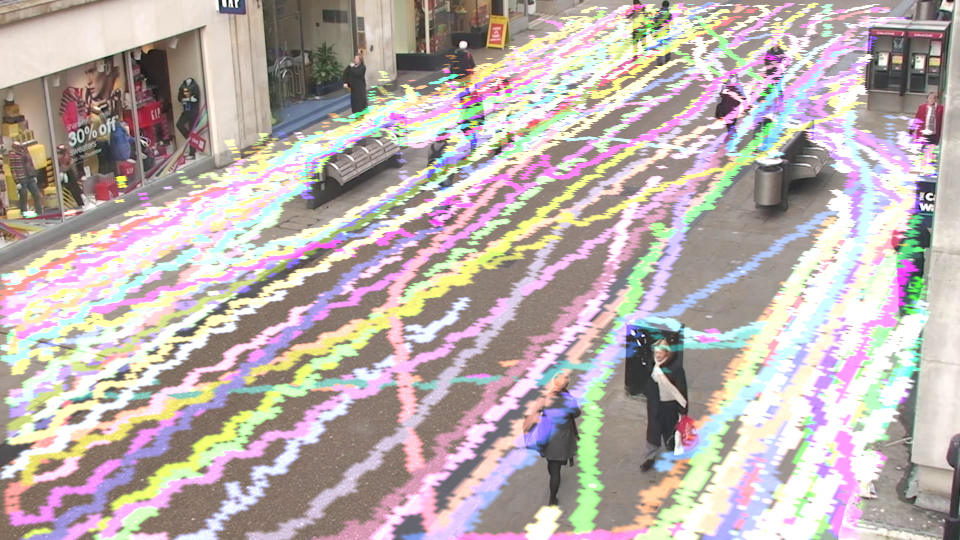}\label{Tracking}}}\hspace{1pt}
\subfloat[Long-term 2D Mosaic heat map of Violations]{
\resizebox*{4.9cm}{!}{\includegraphics{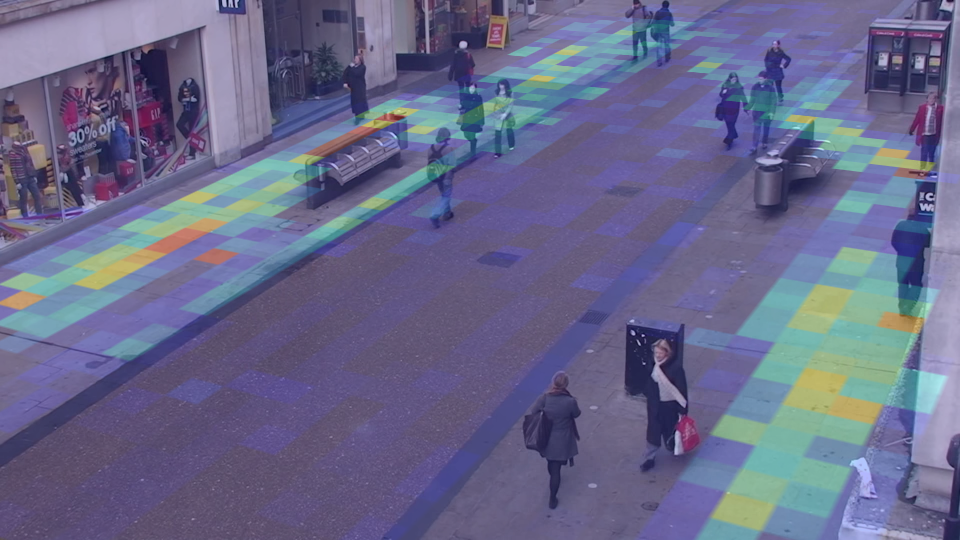}\label{Crowd}}}\hspace{1pt}
\subfloat[2D Mosaic heat map of Tracking + Violations]{
\resizebox*{4.9cm}{!}{\includegraphics{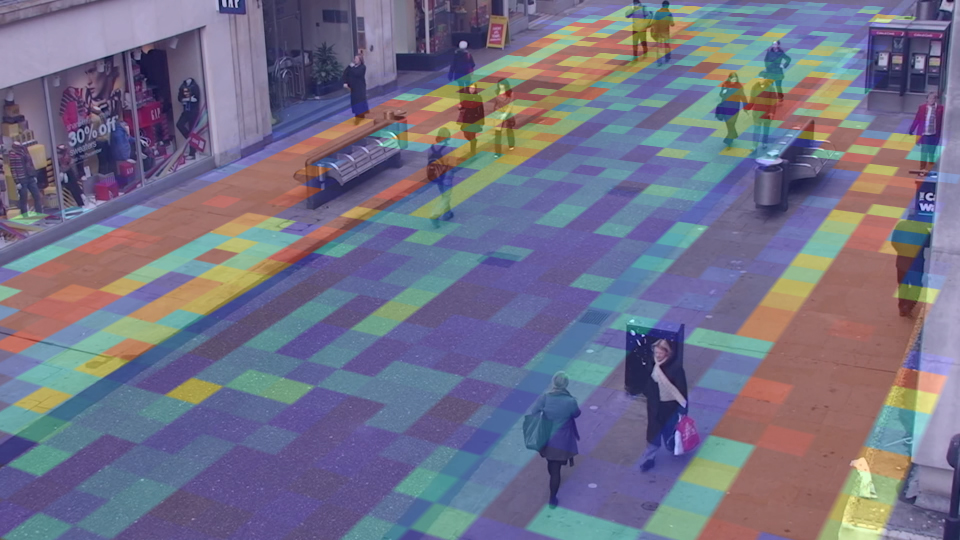}\label{ID}}}

\subfloat[Long-term BEV Tracking heat map]{
\resizebox*{4.3cm}{!}{\includegraphics{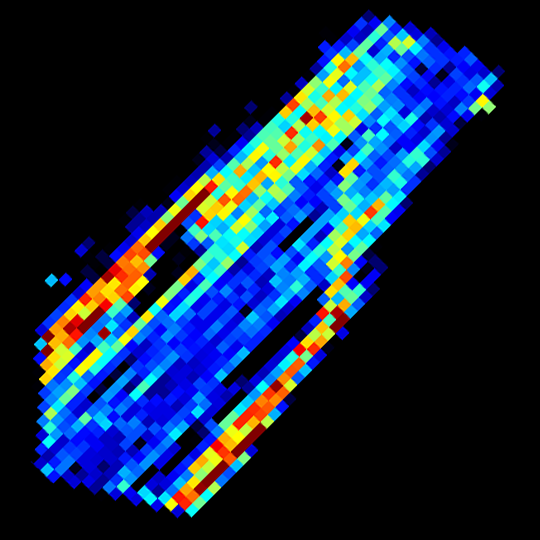}\label{Move-BIRD}}}\hspace{18pt}
\subfloat[Long-term BEV Violation heat map]{
\resizebox*{4.3cm}{!}{\includegraphics{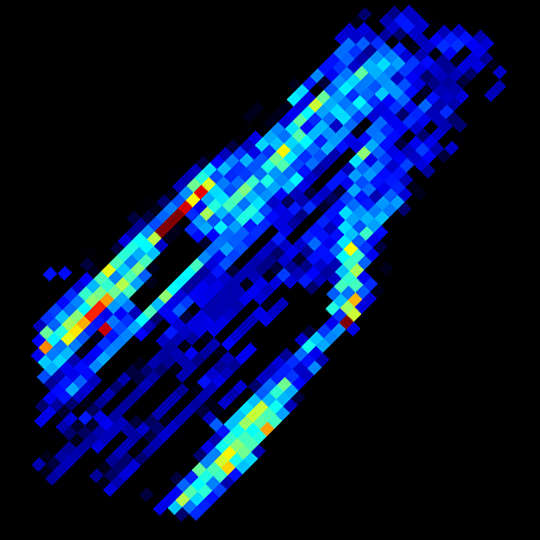}\label{exceptCouples}}}\hspace{18pt}
\subfloat[BEV Tracking + Violation heat map]{
\resizebox*{4.3cm}{!}{\includegraphics{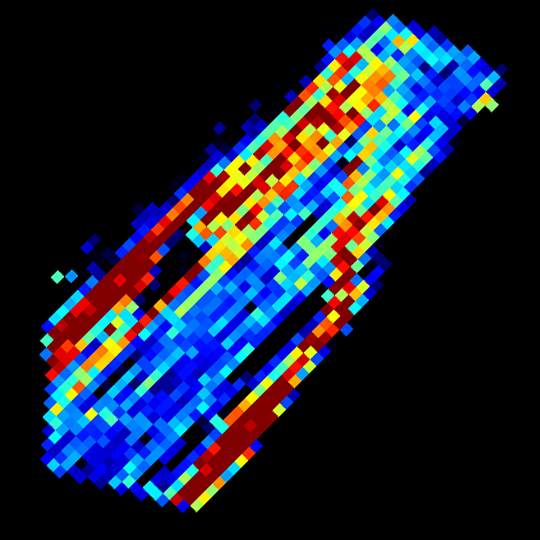}\label{Risk-BIRD}}}

\subfloat[3D Tracking heat map]{
\resizebox*{4.5cm}{!}{\includegraphics{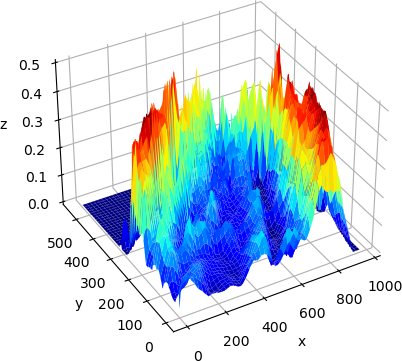}\label{Move-HIST}}}\hspace{18pt}
\subfloat[3D Violation heat map]{
\resizebox*{4.5cm}{!}{\includegraphics{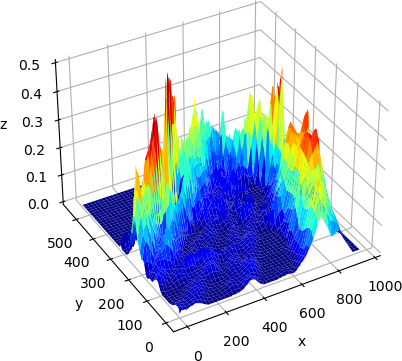}\label{exceptCouples-HIST}}}\hspace{18pt}
\subfloat[3D Tracking + Violation heat map]{
\resizebox*{4.5cm}{!}{\includegraphics{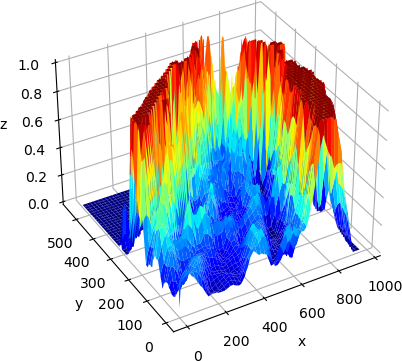}\label{Risk-HIST}}}
\caption{Data analysis based on people detections, tracking, movements, and~breaches of social distancing~measurements.}
\label{violation1}
\end{figure}

Figure~\ref{violation1} shows a visualised output of the discussed experiments including tracking, 2-D and 3-D analytical heat-map of risky zones for the Oxford Town Centre dataset in HSV colour space:
\begin{equation}
\begin{aligned}
R_t = & N(max(G^t , 2 \times S^t), 0, 120) \\
      & R  \in \mathbb{R}^{w \times h}
\end{aligned}
\end{equation}

Figure~\ref{violation1}b shows the tracking paths of the passed people after 2500 frames. 
Figure~\ref{violation1}b illustrates a Mosaic heatmap of people with social distancing violations only.
Figure~\ref{violation1}b shows the summed heatmap of social distancing violations and tracking. 
Figures~\ref{violation1}d--f show the birds-eye view heatmap of long term-tracking, violations, and~the mixed heatmap, respectively. Figures~\ref{violation1}g--i are the corresponding 3D representation of the same heatmaps in the second row, for~better visualisation of safe and risky~zones.

The above configuration was an accumulating approach where all the violations and risky behaviours were added together in order to highlight potentially risky zones in a covered area with poor~ventilation. 

We also thought about cases when there existed a good chance of ventilation where the spread of the virus would not be necessarily accumulative. In~such cases, we considered both increasing and decreasing counts depending on the overall time spent by each individual in each grid cell of the image, as~well as the total absence time of individuals which potentially allowed bringing down the level of~contamination.

\begin{figure}[t!]
\centering
\subfloat[Single-frame 2D crowd map]{
\resizebox*{6.865cm}{!}{\includegraphics{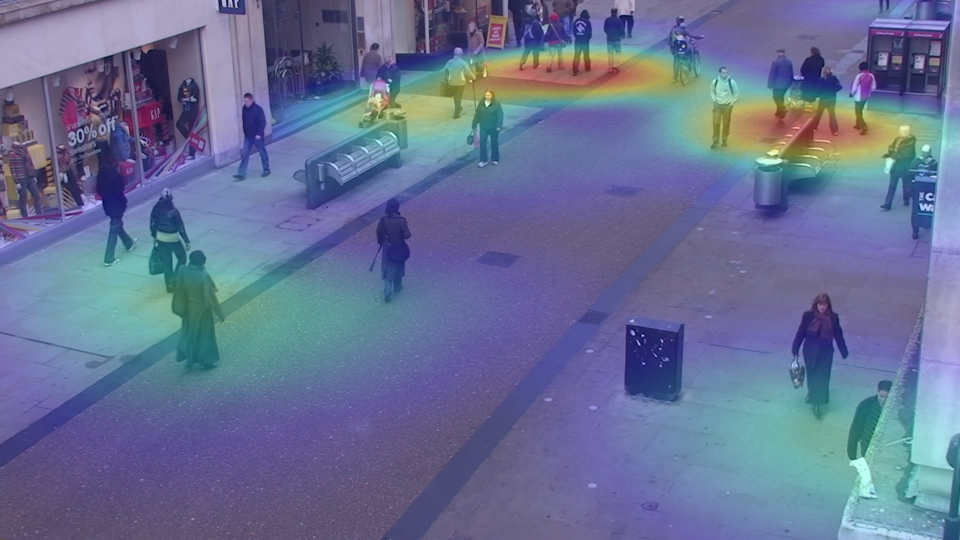}\label{crowd}}}\hspace{1pt}
\subfloat[Single-frame BEV crowd map]{
\resizebox*{3.855cm}{!}{\includegraphics{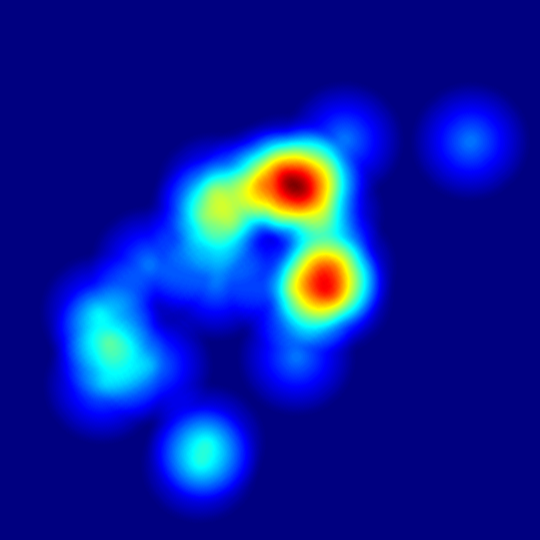}\label{crowdb}}}
\subfloat[Single-frame 3D crowd map]{
\resizebox*{4.5cm}{!}{\includegraphics{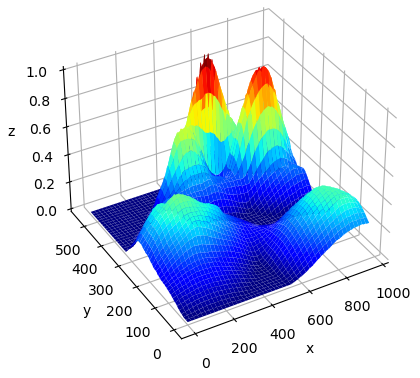}\label{crowdh}}}

\subfloat[Long-term crowd map]{
\resizebox*{6.865cm}{!}{\includegraphics{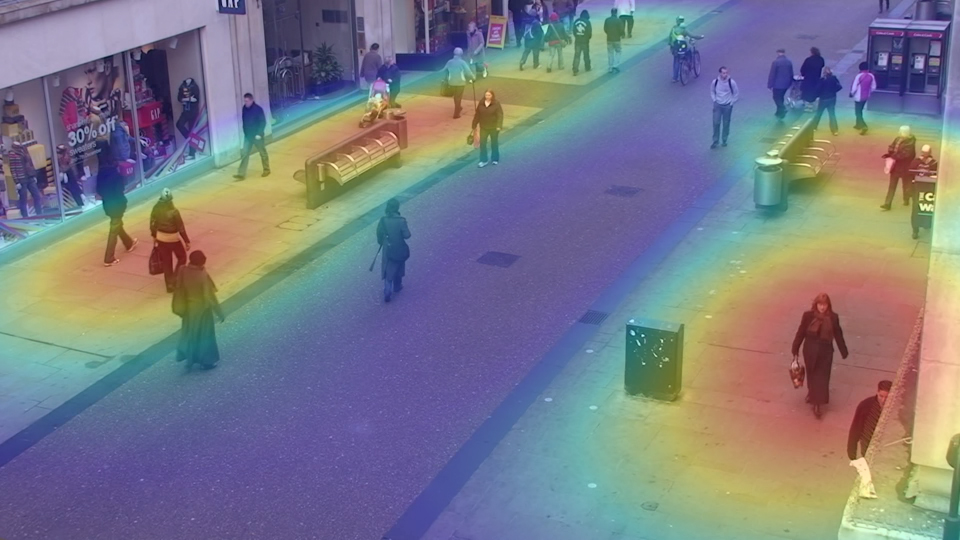}\label{crowdl}}}\hspace{1pt}
\subfloat[Long-term BEV crowd map]{
\resizebox*{3.855cm}{!}{\includegraphics{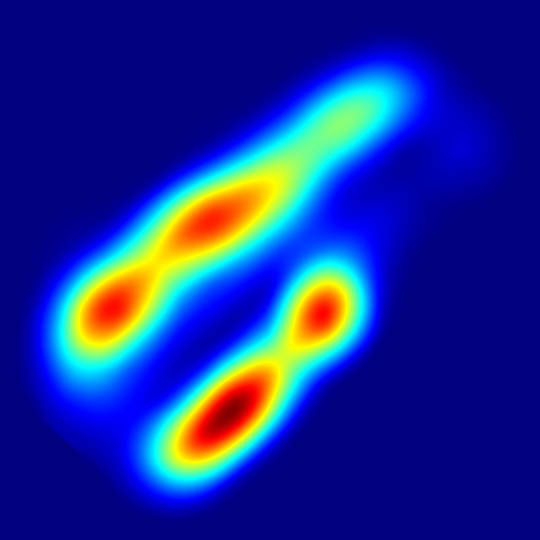}\label{crowdlb}}}
\subfloat[Long-term 3D crowd map]{
\resizebox*{4.5cm}{!}{\includegraphics{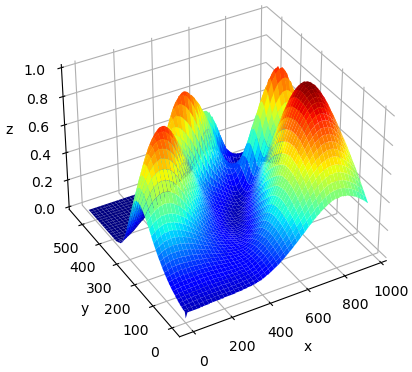}\label{crowdlh}}}
\caption{Single frame vs. Long-term crowd map (2D, BEV, 3D).}
\label{crowd-fig}
\end{figure}

Figure~\ref{crowd-fig} which we named as a crowd map shows the 2D and 3D representation of violations and risk heat maps where we applied both increasing and decreasing contamination trends. The~first row of the image represents a single frame analysis with two peak zones. As~can be seen, those zones belonged to two crowded zones where two large groups of people were walking together and breached the social distancing rule. However, in~other parts of the street a minimal level of risk was identified. This is due to large inter people distances and the consequent time gaps which allowed breathing and therefore a decreasing rate of~contamination.

The second row of Figure~\ref{crowd-fig} shows a long-term crow map which does not necessarily depend on the current frame. This can be a weighted averaging over all of the previous single-frame crow~maps.

One of the extra research questions in Figures~\ref{violation1} and \ref{crowd-fig} is how to define appropriate averaging weights and coefficients $(\alpha, \beta, \gamma)$ in Equation~(\ref{heatmap-f}) and how to normalise the maps over the time. This~is out of the scope of this research and needs further study. 
However, here we aimed at showing the feasibility of considering a diversity of cases using the proposed method with a high level of confidence and accuracy in social distancing monitoring and risk assessment with the help of AI and Computer~Vision. 

\section{Conclusions}\label{conclusion}

We proposed a Deep Neural Network-Based human detector model called DeepSOCIAL to detect and track static and dynamic people in public places in order to monitor social distancing metrics in COVID-19 era and beyond. Various types of state-of-the-art backbones, necks, and~heads were evaluated and investigated. We utilised a CSPDarkNet53 backbone along with an SPP/PAN and SAM neck, YOLO head, Mish activation function. 
We applied the Complete IoU loss function and a Mosaic data augmentation on multi-viewpoint MS COCO and Google Open Image datasets to enrich the training phase, which ultimately led to an efficient and accurate human detector, applicable in various environments using any type of CCTV surveillance~cameras. 

The proposed method was evaluated for Oxford Town Centre dataset, including 7530 frames, and~approximately 150,000 people detection and distance estimation. The~system was able to perform in a variety of challenges including, occlusion, lighting variations, shades, and~partial visibility, and~ proved a major development in terms of accuracy (99.8\%) and speed (24.1 fps) compared to three state-of-the-art techniques. The~system performed real-time using a basic GPU platform or a 10th generation multi-core/multi-thread CPU platform, or~higher.
We adapted an inverse perspective geometric mapping and SORT tracking algorithm for our application to estimate the inter-people distances, and~to track the moving trajectories of the people, infection risk assessment and analysis to the benefit of the health authorities and~governments. 

DeepSOCIAL offered a viewpoint-independent human classification algorithm. Therefore, regardless of the camera angle and position, the~outcome of this research is directly applicable for a wider community of researchers, not only in computer vision, AI, and~health sectors but also in other industrial applications including pedestrian detection for driver assistance systems, autonomous vehicles, anomaly behaviour detections in public and crowd, surveillance security systems, action recognition in sports, shopping centres, public places; and generally, any applications that human detection falls in the centre of attention.\\ 

\noindent \textbf{Supplementary Materials:} 
we publicly share the DeepSOCIAL materials in our \href{https://github.com/DrMahdiRezaei/DeepSOCIAL}{GitHub} repository for the benefit of researchers in the field and research~reproducibility. Sample videos of the project are also accessible via our \href{https://www.youtube.com/channel/UC4KWBb-S3o0QpA-XbxyHUZw}{YouTube} channel and the following links: \href{https://youtu.be/QPZRMWcGuGQ}{Link~1}, \href{https://youtu.be/UHc275F1djk}{Link~2}, \href{https://youtu.be/nPKJKtS-owA}{Link~3}, and \href{https://youtu.be/FwCP2ySDshE}{Link~4}.\\

\reftitle{References}

\end{document}